\documentclass[10pt,twocolumn,letterpaper]{article}

 \usepackage{cvpr} 
\definecolor{cvprblue}{rgb}{0.21,0.49,0.74}
\usepackage[pagebackref,breaklinks,colorlinks,allcolors = cvprblue]{hyperref}

\usepackage{amsmath}
\usepackage{bm}
\usepackage{amsmath}
\usepackage{amssymb}
\usepackage{graphicx}
\usepackage{booktabs}
\usepackage{multirow}
\usepackage{colortbl}
\usepackage{xcolor}
\usepackage{wasysym}
\usepackage{microtype}
\usepackage{algorithm}
\usepackage{algorithmic}
\usepackage{makecell}
\usepackage[switch]{lineno}

\usepackage{newfloat}
\usepackage{listings}

\title{Brain-Inspired Multimodal Spiking Neural Network for Image-Text Retrieval}

\author{
Xintao Zong\textsuperscript{1} \quad
Xian Zhong\textsuperscript{1,}\thanks{Corresponding authors} \quad
Wenxuan Liu\textsuperscript{2,$\ast$} \quad
Jianhao Ding\textsuperscript{2} \quad
Zhaofei Yu\textsuperscript{2} \quad
Tiejun Huang\textsuperscript{2} \\
\textsuperscript{1}Hubei Key Laboratory of Transportation Internet of Things, Wuhan University of Technology \\
\textsuperscript{2}State Key Laboratory for Multimedia Information Processing, Peking University \\
{\tt\small \{zongxt,zhongx\}@whut.edu.cn, \{liuwx66,yuzf12,tjhuang\}@pku.edu.cn, djh01998@alumni.pku.edu.cn}
}

\begin{document}
\maketitle

\begin{abstract}

Spiking neural networks (SNNs) have recently shown strong potential in unimodal visual and textual tasks, yet building a directly trained, low-energy, and high-performance SNN for multimodal applications such as image-text retrieval (ITR) remains highly challenging. Existing artificial neural network (ANN)-based methods often pursue richer unimodal semantics using deeper and more complex architectures, while overlooking cross-modal interaction, retrieval latency, and energy efficiency. To address these limitations, we present a brain-inspired \textbf{C}ross-\textbf{M}odal \textbf{S}pike \textbf{F}usion network (\textbf{CMSF}) and apply it to ITR for the first time. The proposed spike fusion mechanism integrates unimodal features at the spike level, generating enhanced multimodal representations that act as soft supervisory signals to refine unimodal spike embeddings, effectively mitigating semantic loss within CMSF. Despite requiring only two time steps, CMSF achieves top-tier retrieval accuracy, surpassing state-of-the-art ANN counterparts while maintaining exceptionally low energy consumption and high retrieval speed. This work marks a significant step toward multimodal SNNs, offering a brain-inspired framework that unifies temporal dynamics with cross-modal alignment and provides new insights for future spiking-based multimodal research. The code is available at https://github.com/zxt6174/CMSF.


\end{abstract}

\section{Introduction}
\label{sec:intro}

Image-text retrieval (ITR) is a fundamental multimodal task that searches for relevant images given text queries and vice versa. Its importance continues to grow in today’s multimedia-centric world, where \textit{efficient and accurate cross-modal retrieval} underpins commercial applications, information accessibility, and intelligent human-machine interaction.

Existing ITR methods predominantly rely on contrastive learning~\cite{SCAN,VSE-infty,VSRN,USER}, treating paired samples as positives and unpaired samples as negatives. Whether emphasizing local~\cite{SCAN,NAAF} or global~\cite{VSRN,USER} alignment, they share the goal of mapping images and text into a unified high-dimensional semantic space. However, text exhibits strong temporal dependencies across words, and spiking neural networks (SNNs), well known for modeling temporal dynamics, offer a natural fit for capturing such structure. While SNNs excel in energy-efficient, event-driven processing and have achieved success in unimodal tasks~\cite{Zhou_spikformer,LV-SpikeBERT}, their inherent sparsity limits representational richness, hindering fine-grained multimodal understanding. Existing ANN-to-SNN conversion~\cite{youhongANN2SNN} and ANN-guided distillation~\cite{hushengwang} reduce this gap but still require long simulation steps and multi-stage pipelines, making them unsuitable for fast retrieval. Dual-stream architectures~\cite{USER} provide speed but lack deep cross-modal interaction. Thus, a directly trained multimodal SNN that simultaneously offers strong semantic representation and efficient retrieval remains unexplored. 



\begin{figure}[!t]
	\centering
	\includegraphics[width = \linewidth]{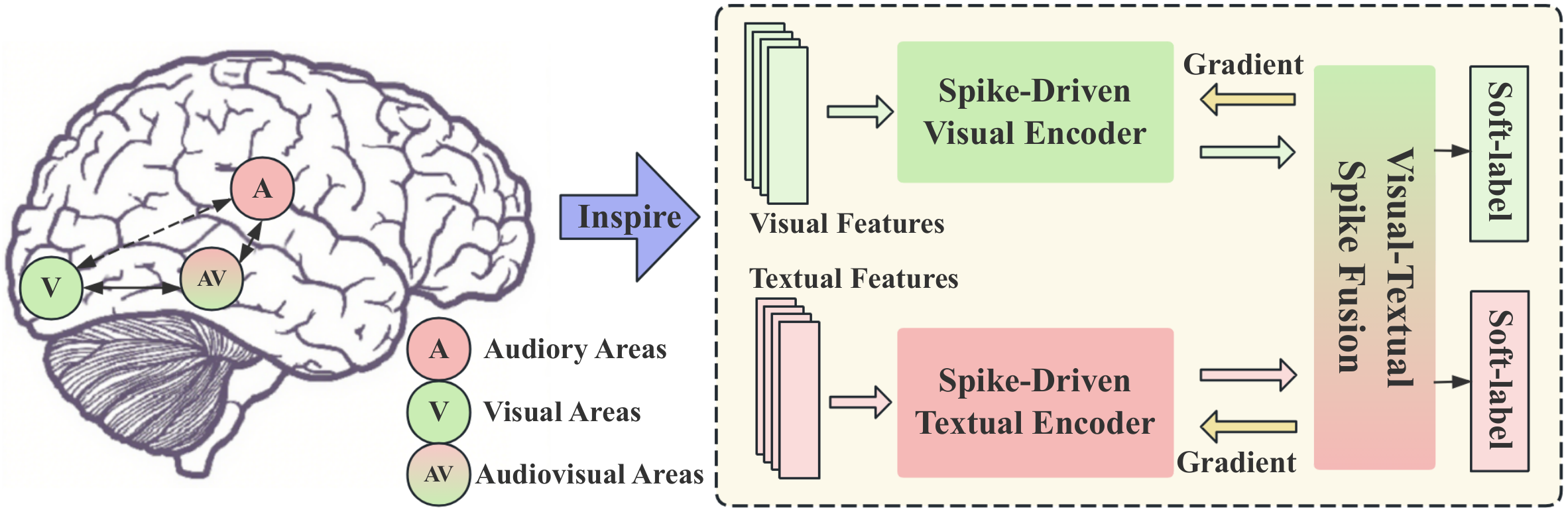} 
	\caption{\textbf{Brain-Inspired Multimodal SNN.} Sensory-specific cortical regions process unimodal information, while higher-order areas such as the audiovisual (AV) cortex integrate multimodal semantics. This hierarchical mechanism inspires our spike-level fusion strategy for constructing a multimodal SNN framework.}
	\label{fig:motivation}
\end{figure}

Cognitive neuroscience~\cite{Sensor} shows that human multimodal cognition involves: 
(1) unimodal encoding and recognition in primary sensory cortices; 
(2) direct neural pathways enabling early cross-modal interaction; and 
(3) higher-order cortical regions responsible for semantic integration. 
We hypothesize that image-text correspondence in ITR follows a similar hierarchical process, motivating a brain-inspired multimodal SNN design.

Building on the biological interpretability and temporal modeling capabilities of SNNs, we propose the \textbf{C}ross-\textbf{M}odal \textbf{S}pike \textbf{F}usion network (\textbf{CMSF}). As illustrated in \cref{fig:motivation}, CMSF, inspired by human cognitive mechanisms, integrates the efficiency of a dual-stream structure with explicit cross-modal interaction through spike-level fusion. The framework comprises two stages:

\textit{(1) Unimodal Spike Embedding:} Floating-point features are encoded into spike embeddings and processed by SNN blocks to capture intra-modal semantics within sparse representations. Early alignment mimics direct neural pathways, temporal pooling forms fine-grained features, and a bidirectional hard-alignment strategy produces similarity matrices.

\textit{(2) Cross-modal Spike Fusion:} We fuse unimodal spike embeddings via a biologically inspired \textbf{Spike Fusion} mechanism that injects complementary information across modalities. The fused spikes generate soft labels that compensate for the information loss inherent to SNNs. Importantly, Spike Fusion is applied only during training, incurring no inference overhead. 

Our main contributions are summarized as threefold:

\begin{itemize}

	\item We propose a brain-inspired multimodal SNN, \textbf{CMSF}, that directly trains a spike-driven architecture for efficient and effective image-text retrieval.
 
	\item We introduce a \textbf{Spike Fusion} mechanism that enriches cross-modal interaction, produces soft supervisory signals, and mitigates the influence of sparsity in SNNs.
 
	\item With minimal time steps, lightweight design, and event-driven efficiency, CMSF achieves top-tier retrieval accuracy and significantly lower energy consumption than state-of-the-art ANN-based models.
 
\end{itemize}

\section{Related Works}
\label{sec:formatting}
\subsection{Image-Text Retrieval}

With the rapid progress of deep learning, image-text retrieval (ITR) has undergone significant advancement. The pioneering SCAN~\cite{SCAN} model employs bottom-up attention~\cite{bottom-up} to identify salient objects and introduces a stacked cross-attention mechanism for similarity computation, inspiring numerous follow-up works~\cite{SCAN-inspire1,SCAN-inspire2,SCAN-inspire3}. Other approaches such as VSE++~\cite{VSE++}, SCO~\cite{SCO}, and CAMP~\cite{CAMP} leverage CNN backbones (\textit{e.g.}, ResNet, VGG) for global feature extraction, while graph convolutional networks have been used to model fine-grained region-word relations~\cite{GCN1,GCN2}.

Further improvements include VSE$\infty$~\cite{VSE-infty}, which adopts a BiGRU-based generalized pooling operator, and USER~\cite{USER}, which incorporates the MoCo~\cite{MoCo} mechanism to expand negative samples under contrastive learning. VSRN++~\cite{VSRN++} enriches representations with BERT~\cite{BERT}, and MMCA~\cite{MMCA} employs Transformer-based architectures to capture both intra- and inter-modal interactions within a unified embedding space. Despite strong retrieval performance, most ANN-based methods pay limited attention to retrieval latency and energy efficiency. Motivated by these limitations, we present a brain-inspired SNN framework tailored for high-efficiency cross-modal retrieval.

\subsection{Spiking Neural Networks}

Early work such as~\cite{Cao_CNN2SNN} explored converting deep CNNs into SNNs by interpreting ANN activations as firing rates. Later studies including~\cite{sengupta-res-snn} improved conversion accuracy through residual architectures and layer-wise weight normalization. Parallel efforts investigated direct training with surrogate gradients: \cite{wu-STBP} introduced spatio-temporal backpropagation (STBP) using approximate derivatives to handle spike non-differentiability, while~\cite{zheng-tdBN} incorporated threshold-dependent batch normalization (tdBN) to enable deeper SNN training.

Following the success of Transformers~\cite{Attention-is-all-you-need}, Spikformer~\cite{Zhou_spikformer} and its variants~\cite{QKformer,Spike-Driven-transformer-V2} proposed a spike-driven self-attention mechanism, marking a milestone in the development of SNNs. In natural language processing, \cite{LV-SpikeCNN-test} proposed a two-step pipeline combining ANN-to-SNN conversion with fine-tuning. Recently, SNNs have achieved notable progress in unimodal computer vision tasks, such as image classification~\cite{ONESTEP}, object detection~\cite{SUYOLO,spikeyolo}, semantic segmentation~\cite{spike2former}, and saliency detection~\cite{liuwenxuan}. However, their application to visual-language cross-modal tasks remains largely unexplored. In this work, we introduce a brain-inspired and directly trained multimodal SNN framework specifically designed for image-text retrieval tasks.


\begin{figure*}[!t] 
	\centering
	\includegraphics[width = \textwidth]{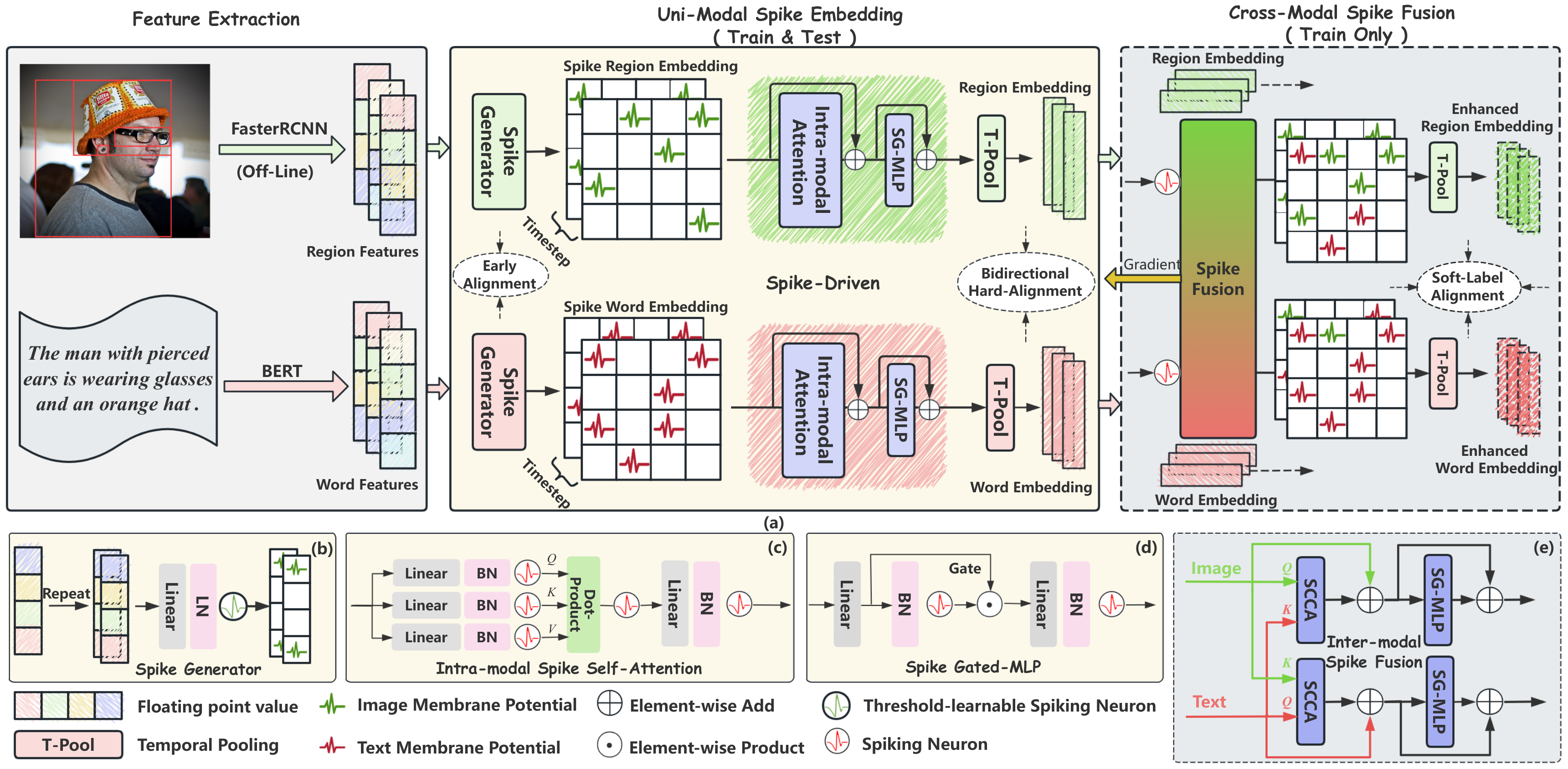} 
	\caption{\textbf{Brain-Inspired Multimodal Spiking Neural Network for Image-Text Retrieval.} (a) Overview of CMSF. (b) Working principle of the Spike Generator. (c,d) Details of Intra-modal Attention blocks. (e) One implementation of the Spike Fusion module. Pre-extracted region and word features are converted into spikes for unimodal semantic modeling within a sparse embedding space. The Spike Fusion module performs spike-level cross-modal interaction, generating enhanced embeddings as soft labels to guide unimodal encoders. This stage is excluded during inference. Bidirectional Hard-Alignment computes the fine-grained similarity matrix, while Early-Alignment and Soft-Label Alignment optimize unimodal encoders from both input and output perspectives. Spike Comb Cross Attention (SCCA) is detailed in \S\ref{subsec:3.5}, with additional fusion variants in the Supplementary Material~A.}
	\label{fig:framework} 
\end{figure*}

\section{Proposed Method} 

An overview of CMSF is shown in \cref{fig:framework}. We first introduce the task preliminaries in \S\ref{subsec:3.1} and the spiking neuron in \S\ref{subsec:3.2}. \S\ref{subsec:3.3} presents the feature extraction method, which can be flexibly replaced with other feature extractors, \S\ref{subsec:3.4} describes spike-driven intra-modal modeling and the bi-directional hard alignment mechanism, and \S\ref{subsec:3.5} details the cross-modal spike fusion process and soft-label alignment strategy. Finally, \S\ref{subsec:3.6} presents the alignment objective function. 

\subsection{Preliminary Definition} \label{subsec:3.1} 

To leverage the temporal dynamics of SNNs, we align image-text data with discrete SNN time steps. Text sequences can be treated as temporal data~\cite{SeqSNN}, where each text step $\Delta T$ is divided into $T_{s}$ sub-steps, allowing neuron activation at each sub-step, \textit{i.e.}, $\Delta T = T_{s}\Delta t$. This bridges word steps $\Delta T$ with discrete SNN time steps $\Delta t$. Similarly, region features are processed sequentially.

\subsection{Spiking Neuron} \label{subsec:3.2} 

As the fundamental unit of SNNs, a spiking neuron receives input currents $X[t]$ and accumulates membrane potentials, which are compared to a threshold to determine firing. The dynamics of the leaky integrate-and-fire (LIF) neuron model~\cite{third-network-LIF} are: 
\begin{align} 
	H[t] & = V[t-1] + \frac{1}{\tau} \left(X[t] - \left(V[t-1]-V_\mathrm{reset} \right) \right), \\ 
	S[t] & = \begin{cases} 1, & \text{if }H[t] \ge V_{\mathrm{th}}, \\ 0, & \text{otherwise}, \end{cases} \\ 
	V[t] & = H[t] \left(1 - S[t] \right) + V_\mathrm{reset}S[t], 
\end{align}
where $\tau$ is the membrane time constant. When $H[t]$ exceeds the threshold $V_{\mathrm{th}}$, the neuron emits a spike $S[t]$; if spiking occurs, $V[t]$ resets to $V_\mathrm{reset}$, otherwise it remains $H[t]$.

\subsection{Feature Extraction}
\label{subsec:3.3} 


In ITR tasks, pre-extracting region features from images is a widely adopted paradigm~\cite{SCAN,VSE++,VSE-infty}, allowing subsequent alignment networks to focus on efficiency and accuracy. The feature extraction module is decoupled; a better extraction backbone will bring better performance. For a fair comparison, we follow the same practice as ANN baselines~\cite{MaxMatch,VSE-infty,HREM,CHAN,USER} to pre-extract floating-point features. 


\vspace{-8pt}
\paragraph{Region Features.} 
Following prior work~\cite{SCAN,VSE++,VSE-infty}, we extract $N$ region features $R_o = \{r_1,\dots,r_N\} \in \mathbb{R}^{N \times 2048}$ using Faster R-CNN~\cite{FasterRCNN} in a top-down manner~\cite{bottom-up}, where $N$ denotes the number of detected regions.

\vspace{-8pt}
\paragraph{Word Features.}
We obtain $L$ word-level features $E_o = \{e_1,\dots,e_L\} \in \mathbb{R}^{L \times 768}$ from the final layer of a pre-trained sequential model, BERT~\cite{BERT}. 

Then we add a fully connected layer to map each region and word to a common embedding dimension $D$, producing floating-point features $R_f \in \mathbb{R}^{N \times D}$ and $E_f \in \mathbb{R}^{L \times D}$.

\subsection{Unimodal Spike Embedding}
\label{subsec:3.4}

We adopt a dual-stream symmetric architecture for unimodal spike embedding, denoting regions $R$ and words $E$ by $X$ in the following formulations.

\vspace{-8pt}
\paragraph{Spike Generator.} 
To meet the spatiotemporal requirements of SNNs, we employ a Spike Generator to convert floating-point region and word features into spiking pattern embeddings:
\begin{align}
	X_{s} = \mathcal{TLSN} \left(\mathrm{LN} \left(\mathrm{Repeat} \left(X_{f},T \right) \right) \right),
\end{align}
where $\mathrm{Repeat}(X_{f},T)$ duplicates features $T$ times, LN denotes layer normalization, and $\mathcal{TLSN}$ is a threshold-learnable spiking neuron. This produces spike region embeddings $R_s \in \mathbb{R}^{T \times N \times D}$ and spike word embeddings $E_s \in \mathbb{R}^{T \times L \times D}$, catering to the following spiking layers for event-driven modeling.

\vspace{-8pt}
\paragraph{Intra-Modal Attention.} 
The sparse 0-1 spikes generated by the Spike Generator inevitably lead to information loss compared with the original floating-point representations. Therefore, it is crucial to design a biologically inspired, spike-driven, and high-performing SNN framework to model intra-modal semantic relations within this sparse embedding space. 

The spike self-attention (SSA) mechanism~\cite{Zhou_spikformer}, biologically inspired and competitive with vanilla self-attention in unimodal classification tasks, is instantiated as an optimized \{SSA+SGMLP\} block within our symmetric dual-stream architecture to model intra-modal semantic representations. The calculation process of SSA is formulated as follows:
\begin{gather}
	Q_S ,K_S,V_S = \mathcal{SN}_I\left(\mathrm{BN}\left(X_s W_I \right)\right), I \in (Q,K,V),\\
	A_S = \mathcal{SN}\left(\mathrm{BN}\left(Q_S K_S^TV_S*s\right)\right), \\
	\mathrm{SSA} \left(Q_S,K_S,V_S \right) = \mathcal{SN}\left(\mathrm{BN}\left(A_SW_{A}\right)\right),
\end{gather}
where $W_Q$, $W_K$, $W_V$, and $W_A$ are learnable matrices of linear layers, and $s$ is a scaling factor.

\vspace{-8pt}
\paragraph{Spike Gated-MLP.} 
Spike signals in SNNs often attenuate across layers~\cite{ONESTEP}, hindering effective information propagation. We replace standard MLP modules in each attention block with a Spike Gated-MLP (SG-MLP), which introduces a gating mechanism~\cite{GRNN,gmlp}, implemented as controllable inhibitory or excitatory synapses~\cite{excitation-inhibition}, to preserve pre-activation information and selectively control spike trains:
\begin{align}
	G_S & = \mathcal{SN} \left(X_s W_G \right), \\ 
	P_F & = X_s W_P, \\ 
	\mathrm{SGMLP} \left(G_S,P_F \right) & = \mathcal{SN} \left( \left(G_S \odot P_F \right) W_O \right),
\end{align}
where $W_G$, $W_P$, and $W_O$ are learnable matrices, and $\odot$ denotes element-wise multiplication.

\vspace{-8pt}
\paragraph{Temporal Pooling.} 
Outputs of the Intra-modal Attention blocks are aggregated along the temporal dimension via a weighted average, where learnable weights $w_t$ dynamically attend to the importance of each time step $t$, producing a floating-point representation $X_s$ for similarity computation:
\begin{align}
	\mathrm{TPool} \left(X_s \right) = \sum_{t = 1}^T w_{t} X_{s} \left(t \right).
\end{align}

\vspace{-8pt}
\paragraph{Pipeline Summary.}

In summary, the Unimodal Spike Embedding stage processes input image-text data as:
\begin{align}
	X_f & = \mathrm{Linear} \left(X_o \right), \\ 
	X_s & = \mathrm{SpikeGenerator} \left(X_f,T \right) , \\ 
	X_s' & = X_s + \mathrm{SSA} \left(X_s \right), \\ 
	X_s'' & = X_s' + \mathrm{SGMLP} \left(X_s' \right), \\ 
	\tilde{X} & = \mathrm{TPool} \left(X_s'' \right), 
	\label{eq:78th}
\end{align}
where $\mathrm{TPool}(\cdot)$ denotes temporal weighted pooling for aggregating temporal information, and $\tilde{X} \in \mathbb{R}^{K \times D}$, $X_{s}'' \in \mathbb{R}^{T \times K \times D}$, with $K$ corresponding to the number of regions $N$ (for image data) or words $L$ (for text data).

\begin{figure}[!t]
	\centering
	\includegraphics[width = \linewidth]{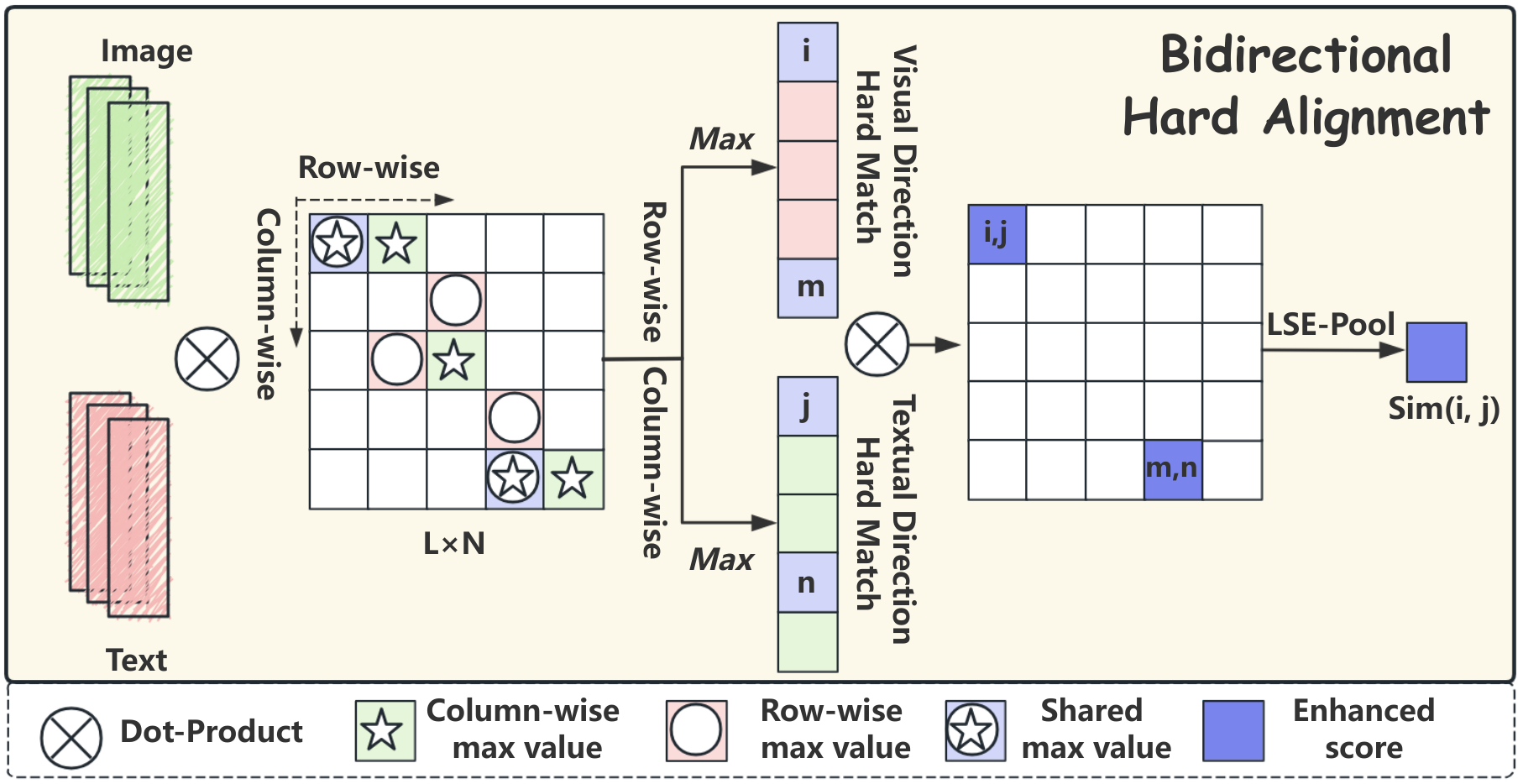} 
	\caption{\textbf{Illustration of Our Bidirectional Hard Alignment.} It identifies fine-grained region-word hard matches and integrates them to compute the overall image-text similarity.}
	\label{fig:BHA}
\end{figure}

\vspace{-8pt}
\paragraph{Bidirectional Hard Alignment.} 
We compute the fine-grained similarity tensor 
$s(E,R) \in \mathbb{R}^{B \times B \times N \times L}$, where $B$ denotes the batch size. 
Following CHAN~\cite{CHAN}, for each word we select the maximum similarity across all regions:
\begin{align}
	s \left(e_i,R \right) = \max_{j = 1,\dots,N} \left(s_{ij} \right) \in \mathbb{R}^{B \times B \times L}.
\end{align}
However, this ignores the reverse direction, where each region can match its most relevant word:
\begin{align}
	s \left(E,r_j \right) = \max_{i = 1,\dots,L} \left(s_{ij} \right) \in \mathbb{R}^{B \times B \times N}.
\end{align}
Due to inconsistencies in image-text data, these two maxima rarely coincide in $s(E,R)$. As shown in \cref{fig:BHA}, we address this by computing their outer product to form an enhanced similarity matrix:
\begin{align}
	\overline{s} \left(E,R \right) = s \left(e_i,R \right) \otimes s \left(E,r_j \right) \in \mathbb{R}^{B \times B \times N \times L},
\end{align}
where $\otimes$ denotes the outer product. This operation amplifies the similarity of relevant region-word pairs $(e_i, r_j)$, boosting $\overline{s}_{ij}$. Finally, we apply 2D LogSumExp pooling to obtain the global similarity matrix:
\begin{align}
	S \left(E,R \right) = \frac{1}{\alpha}\log \left(\sum_{n = 1}^N\sum_{l = 1}^L \exp \left(\alpha \overline{s}_{ij} \right) \right)
	\in \mathbb{R}^{B \times B},
	\label{Sim}
\end{align}
where $\alpha$ controls pooling smoothness.

\subsection{Cross-Modal Spike Fusion}
\label{subsec:3.5}

Our Spike Fusion module selectively activates salient membrane potentials in one modality, preserves cross-modal interactions and retains shared spike activation information, and suppresses redundant spikes as mutual noise. It is implemented through element-wise multiplication of binary matrices across modality embeddings, functioning as a biologically inspired spike-driven cross-attention mechanism.

\begin{figure}[!t]
	\centering
	\includegraphics[width = \linewidth]{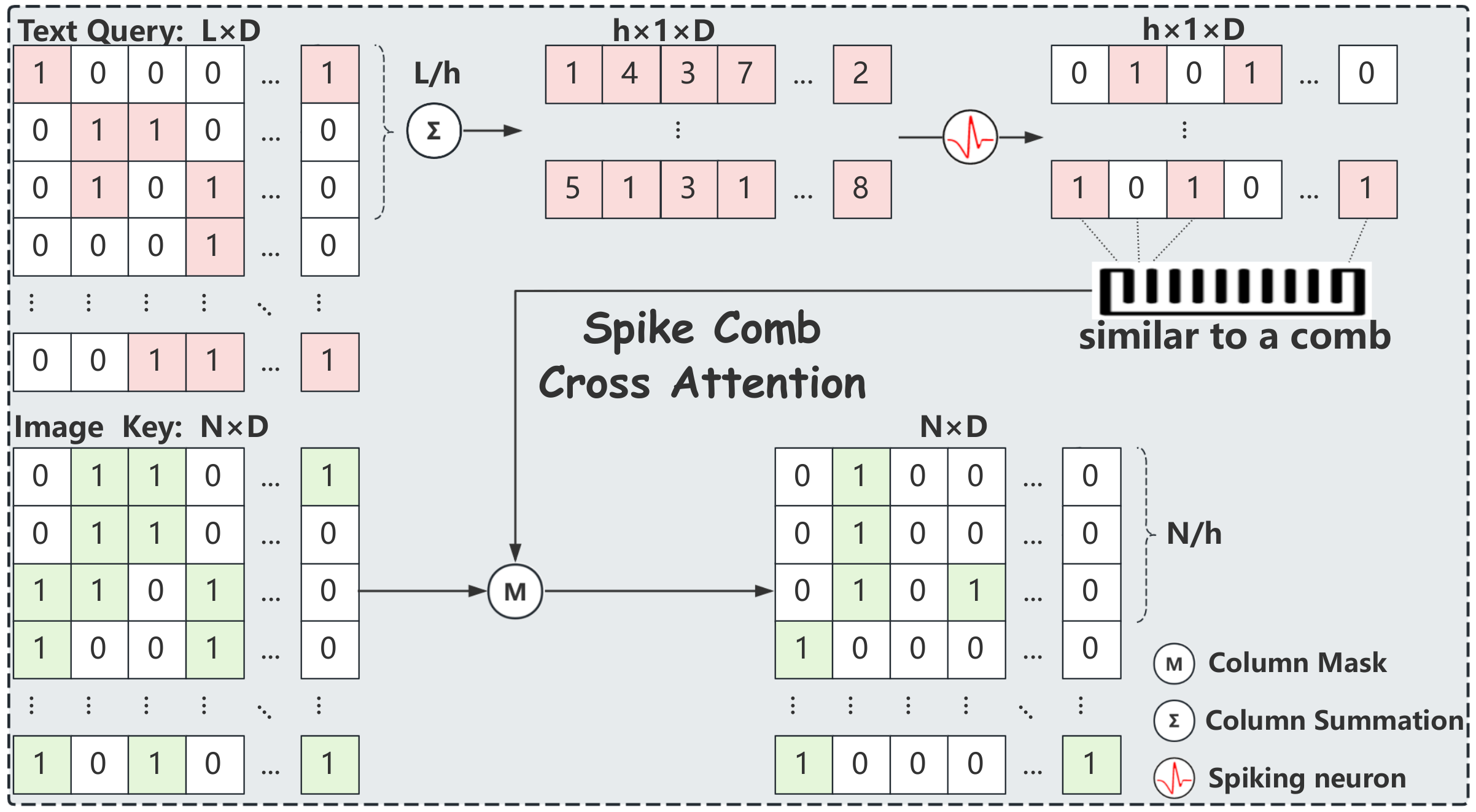} 
	\caption{\textbf{Details of Our Spike Comb Cross Attention Structure.} With the mask operation, it achieves lower time complexity and reduced energy consumption.}
	\label{fig:spike-comb-cross-attention}
\end{figure}

\vspace{-8pt}
\paragraph{Spike Comb Cross Attention.} 
To reduce computational load and training complexity on large datasets, we propose a task-specific cross-attention mechanism inspired by QKFormer~\cite{QKformer}, termed Spike Comb Cross Attention (SCCA), as shown in \cref{fig:spike-comb-cross-attention}. Given text spike embeddings $Q \in \mathbb{R}^{L \times D}$ and image spike embeddings $K \in \mathbb{R}^{N \times D}$, we divide $Q$ into $h$ heads of shape $(L/h, D)$, sum within each head’s spatial dimension, and pass the result through a LIF neuron to obtain $h$ ``combs'':
\begin{align}
	\mathrm{Comb}_i = \mathcal{SN} \left(\sum_{j = 1}^{L/h} Q_{i,j}\right), \quad i = 1,\dots,h.
	\label{Comb}
\end{align}
Each comb’s ``teeth'' capture activation frequencies at different embedding positions. These combs are filtered against $K$ along its dimension $N$, discarding unmatched membrane potentials and retaining aligned activations to ensure spike distribution consistency across modalities:
\begin{align}
	X' = \mathrm{Comb} \odot K,
\end{align}
where $\odot$ denotes duplicating each comb $Comb_i \in \mathbb{R}^{1 \times D}$ into a tensor of shape $(N/h) \times D$, followed by element-wise multiplication with $K$.


For multi-timestep inputs $E \in \mathbb{R}^{T \times L \times D}$ and $R \in \mathbb{R}^{T \times N \times D}$, SCCA achieves complementary alignment across multiple dimensions. Along the temporal dimension $T$, neurons are synchronized by firing intervals; along semantic spatial dimensions $L$ and $N$, dividing inputs into $h$ combs realizes block-level region-word alignment; along the embedding dimension $D$, multiple ``teeth'' act as dynamic masks, aligning both spike counts and neuronal topology. Consequently, our Spike Fusion module generates high-quality, information-enhanced soft labels.

We further design alternative spike-driven fusion variants, including Spike Cross Attention and Spike Concat-Attention. Details are provided in the \underline{Supplementary Material~A}.

\vspace{-8pt}
\paragraph{Spike Fusion Soft-Label Alignment.} 
Unlike teacher-based~\cite{CUSA} or self-distillation~\cite{SoftCLIP} methods, our fusion directly merges multimodal spike embeddings, emulating higher-order cortical integration to generate high-quality soft labels. These labels align shallow unimodal features with deeper cross-modal representations, injecting complementary information and mitigating information loss in SNN-based encoders.

\begin{figure}[!t]
	\centering
	\includegraphics[width = \linewidth]{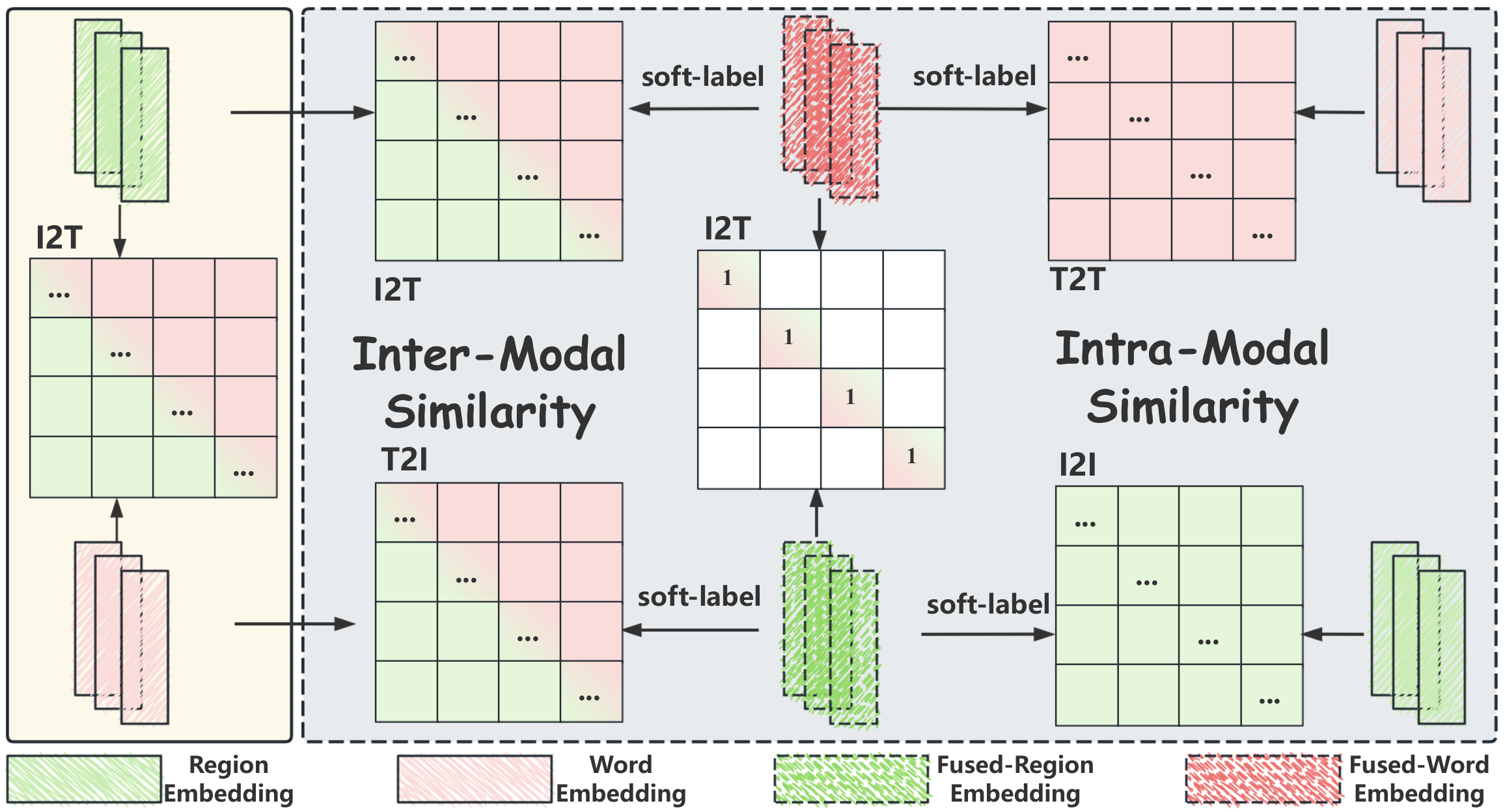} 
	\caption{\textbf{Illustration of Spike Fusion Soft-Label Alignment Strategy.} The left part represents the similarity computation involved in both training and inference; the right part is training-only.}
	\label{fig:soft}
\end{figure}

At this stage, unimodal outputs $\tilde{R}$ and $\tilde{E}$ are fused into $\overline{R}$ and $\overline{E}$. As shown in \cref{fig:soft}, we compute the contrastive loss on $(\overline{R}, \overline{E})$ and use them as soft labels to guide the unimodal encoders, encouraging $\tilde{R}$ and $\tilde{E}$ to align accordingly. The alignment strategy includes:
\textit{1) Inter-modal similarity alignment}, computing similarity matrices $S(\tilde{E}, \tilde{R})$, $S(\overline{E}, \overline{R})$, $S(\tilde{E}, \overline{R})$, and $S(\overline{E}, \tilde{R})$ following \cref{Sim}; and 
\textit{2) Intra-modal similarity alignment}, computing $S(\tilde{E}, \overline{E})$ and $S(\tilde{R}, \overline{R})$ following \cref{Sim} similarly.

\vspace{-8pt}
\paragraph{Early Alignment.} 
To mimic direct neural pathways between sensory-specific areas that enable early cross-modal interaction while avoiding data leakage in the dual-stream design, we introduce an early alignment mechanism. Specifically, we compute $S(E_f, R_f)$ via \cref{Sim} and apply a loss function to enforce initial similarity between floating-point modality features, preventing spike discrepancies from being amplified during propagation. \S\ref{subsec:4.3} details the necessity of this approach in the dual-stream SNN.

\subsection{Alignment Objective}
\label{subsec:3.6}

Given a batch of $B$ matched image-text pairs, we first compute the global similarity matrix $S \in \mathbb{R}^{B \times B}$ using our Bidirectional Hard Alignment in \cref{Sim}, where each element $S_{ij}$ denotes the similarity between the $i$-th image and the $j$-th text. We then apply the InfoNCE loss. The image-to-text loss is defined as:
\begin{align}
	\mathcal{L}_{\mathrm{i2t}} = \frac{1}{B}\sum_{i = 1}^B \log \sum_{j\neq i} \exp \left(\frac{1}{\tau} \left(S_{ij}-S_{ii} \right) \right),
\end{align}
where $\tau$ is a temperature hyperparameter controlling the distribution sharpness. Similarly, the text-to-image loss is defined as:
\begin{align}
	\mathcal{L}_{\mathrm{t2i}} = \frac{1}{B}\sum_{i = 1}^B \log \sum_{j\neq i} \exp \left(\frac{1}{\tau} \left(S_{ji}-S_{ii} \right) \right).
	\label{losst2i}
\end{align}
The overall pairwise loss between image representations $R$ and text representations $E$ is:
\begin{align}
	\mathcal{L}(E,R) = \frac{1}{2}\left(\mathcal{L}_{\mathrm{i2t}}+\mathcal{L}_{\mathrm{t2i}}\right).
	\label{loss}
\end{align}

We denote the initial floating-point representations extracted in the first stage as $R_f$ and $E_f$, the unimodal embeddings after SNN processing as $\tilde{R}$ and $\tilde{E}$, and the enhanced representations obtained through multimodal Spike Fusion as $\overline{R}$ and $\overline{E}$. Based on the similarity matrices obtained in Spike Fusion Soft-Label Alignment and \cref{loss}, we define:
\begin{align*}
	& \mathcal{L}_{\mathrm{Early}} = \mathcal{L}(E_f, R_f) & \text{(early-stage alignment)}, \\ 
	& \mathcal{L}_{\mathrm{Basic}} = \mathcal{L}(\tilde{E}, \tilde{R}) & \text{(dual-stream alignment)}, \\ 
	& \mathcal{L}_{\mathrm{Fusion}} = \mathcal{L}(\overline{E}, \overline{R}) & \text{(fusion-level alignment)}, \\ 
	& \mathcal{L}_{\mathrm{Inter}} = \mathcal{L}(\tilde{E}, \overline{R}) + \mathcal{L}(\overline{E}, \tilde{R}) & \text{(soft-label alignment)}, \\ 
	& \mathcal{L}_{\mathrm{Intra}} = \mathcal{L}(\tilde{E}, \overline{E}) + \mathcal{L}(\tilde{R}, \overline{R}) & \text{(soft-label alignment)}.
\end{align*}

Following the plug-and-play alignment strategy in CUSA~\cite{CUSA} and the cross-modal interaction loss design used in HREM~\cite{HREM}, we define:
\begin{align}
	\mathcal{L}_{\mathrm{Late}} = \mathcal{L}_{\mathrm{Basic}} + \mathcal{L}_{\mathrm{Fusion}} + \mathcal{L}_{\mathrm{Inter}} + \mathcal{L}_{\mathrm{Intra}}.
\end{align}
Finally, our total training objective is formulated as:
\begin{align}
	\mathcal{L}_{\mathrm{Total}} = \lambda \mathcal{L}_{\mathrm{Early}} + \left(1- \lambda \right)\mathcal{L}_{\mathrm{Late}},
	\label{losstotal}
\end{align}
where $\lambda$ is the loss weight, which balances the contributions of different loss components.

\section{Experimental Results}

\subsection{Experimental Details}

\paragraph{Datasets.} 
We evaluate CMSF on two widely used benchmarks: \textsc{Flickr30K}~\cite{Flickr30K} and \textsc{MSCOCO}~\cite{MSCOCO}. \textsc{Flickr30K} contains 31,783 images and \textsc{MSCOCO} includes 123,287 images, each paired with five captions. Following the standard evaluation protocol~\cite{SCAN,USER}, \textsc{Flickr30K} is split into 29,000/1,000/1,000 images for training/validation/testing, while \textsc{MSCOCO} uses 113,287/1,000/5,000 images.

\vspace{-8pt}
\paragraph{Metrics.} 
Retrieval performance is evaluated with Recall@$K$ ($K = 1,5,10$), measuring the percentage of queries whose ground-truth match is ranked within the top-$K$. We additionally report R@Sum, the sum of all six Recall@$K$ values. 


\vspace{-8pt}
\paragraph{Implementation Details.} 
CMSF is implemented with the SpikingJelly~\cite{Spikingjelly} framework and trained on a single NVIDIA RTX 4090 GPU (24 GB memory). We set the embedding dimension to $D = 1024$, batch size $B=160$, $\alpha = 0.1$ in \cref{Sim}, the number of teeth $h = 6$ in \cref{Comb}, the loss smoothing factor $\gamma = 0.01$ in \cref{losst2i}, and $\lambda = 0.5$ in \cref{losstotal}. Training is performed for 35 epochs with an initial learning rate of $5\times 10^{-4}$, decayed by 10× during the final 15 epochs. For Intra-modal Attention and Spike Generators, we use $T = 2$ time steps, 1 layer depth, and AdamW with a learning rate of $5\times 10^{-4}$. The pretrained BERT encoder~\cite{BERT} is fine-tuned with $5\times 10^{-5}$ for stability. Spike Fusion also uses $T = 2$ and 1 layer depth but employs a larger learning rate ($5\times 10^{-3}$) for faster convergence. 



\begin{table*}[!t]
	\centering
	\footnotesize
	\setlength{\tabcolsep}{5pt}
	\caption{\textbf{Results on \textsc{MSCOCO 1K} and \textsc{Flickr30K 1K} Test Sets.} Best results are in \textbf{bold}, and the second-best scores are \underline{underlined}. $\ast$ indicates ensemble results. Additional results on pretrained VLMs (\textit{e.g.}, CLIP~\cite{CLIP}) are reported in Supplementary Material~B.}
	\begin{tabular}{l|ccc|ccc|c|ccc|ccc|c}
	\toprule[1.1pt]
	\multirow{3}[4]{*}{Methods} & \multicolumn{7}{c|}{\textsc{MSCOCO} 1K Test Set} & \multicolumn{7}{c}{\textsc{Flickr30K} 1K Test Set} \\ 
	\cmidrule(lr){2-8} \cmidrule(lr){9-15}
	& \multicolumn{3}{c|}{Image-to-Text} & \multicolumn{3}{c|}{Text-to-Image} & \multirow{2}[2]{*}{R@Sum}
	& \multicolumn{3}{c|}{Image-to-Text} & \multicolumn{3}{c|}{Text-to-Image} & \multirow{2}[2]{*}{R@Sum} \\ 
	\cmidrule(lr){2-4} \cmidrule(lr){5-7} \cmidrule(lr){9-11} \cmidrule(lr){12-14}
	& R@1 & R@5 & R@10 & R@1 & R@5 & R@10 &
	& R@1 & R@5 & R@10 & R@1 & R@5 & R@10 & \\ 
	\midrule
	\multicolumn{15}{c}{\textit{Faster R-CNN + BiGRU}} \\ 
	\midrule
	SCAN$\ast$~\cite{SCAN} & 72.7 & 94.8 & 98.4 & 58.8 & 88.4 & 94.8 & 507.9 & 67.4 & 90.3 & 95.8 & 48.6 & 77.7 & 85.2 & 465.0 \\ 
	VSRN$\ast$~\cite{VSRN} & 76.2 & 94.8 & 98.2 & 62.8 & 89.7 & 95.1 & 516.8 & 71.3 & 90.6 & 96.0 & 54.7 & 81.8 & 88.2 & 482.6 \\ 
	VSE$\infty$~\cite{VSE-infty} & 78.5 & 96.0 & 98.7 & 61.7 & 90.3 & 95.6 & 520.8 & 76.5 & 94.2 & 97.7 & 56.4 & 83.4 & 89.9 & 498.1 \\ 
	NAAF~\cite{NAAF} & 78.1 & 96.1 & 98.6 & 63.5 & 89.6 & 95.3 & 521.2 & 79.6 & 96.3 & 98.3 & 59.3 & 83.9 & 90.2 & 507.6 \\ 
	\rowcolor{gray!20}
	CMSF(Ours) & 78.9 & 96.2 & 98.7	& 63.6	& 91.0	& 96.3	& 524.7 & 80.7	& 95.0	& 97.6	& 61.3	& 85.9	& 91.3	& 511.8\\
	\midrule
	\multicolumn{15}{c}{\textit{Faster R-CNN + BERT}} \\ 
	\midrule
	MMCA~\cite{MMCA} & 74.8 & 95.6 & 97.7 & 61.6 & 89.8 & 95.2 & 514.7 & 74.2 & 92.8 & 96.4 & 54.8 & 81.4 & 87.8 & 487.4 \\ 
	TERAN$\ast$~\cite{TERAN} & 80.2 & 96.6 & \textbf{99.0} & \underline{67.0} & \underline{92.2} & \textbf{96.9} & 531.9 & 79.2 & 94.4 & 96.8 & 63.1 & 87.3 & 92.6 & 513.4 \\ 
	VSE$\infty$~\cite{VSE-infty} & 79.7 & 96.4 & 98.9 & 64.8 & 91.4 & 96.3 & 527.5 & 81.7 & 95.4 & 97.6 & 61.4 & 85.9 & 91.5 & 513.5 \\ 
	VSRN++$\ast$~\cite{VSRN++} & 77.9 & 96.0 & 98.5 & 64.1 & 91.0 & 96.1 & 523.6 & 79.2 & 94.6 & 97.5 & 60.6 & 85.6 & 91.4 & 508.9 \\ 
	CHAN~\cite{CHAN} & 81.4 & \textbf{96.9} & 98.9 & 66.5 & 92.1 & 96.7 & \underline{532.6} & 80.6 & 96.1 & 97.8 & \underline{63.9} & \underline{87.5} & \underline{92.6} & 518.5 \\ 
	HREM~\cite{HREM} & 81.1 & 96.6 & 98.9 & 66.1 & 91.6 & 96.5 & 530.7 & \underline{83.3} & 96.0 & \underline{98.1} & 63.5 & 87.1 & 92.4 & 520.4 \\ 
	USER~\cite{USER} & 
\underline{82.8} & \underline{96.8} & 98.8 & 66.1 & 90.6 & 95.6 & 530.5 & 82.7 & \textbf{97.0} & \textbf{98.3} & 63.1 & 86.7 & 92.1 & 519.9 \\ 
	MaxMatch~\cite{MaxMatch} & \textbf{83.0} & \textbf{96.9} & 98.9 & 66.4 & 91.9 & 96.6 & \textbf{533.8} & \textbf{84.2} & 96.1 & 97.9 & 63.2 & 87.3 & 92.2 & \underline{520.8} \\ 
	\rowcolor{gray!20}
	{CMSF (Ours)} & 81.9 & 96.7 & \underline{98.9} & \textbf{67.1} & \textbf{92.4} & \underline{96.8} & \textbf{533.8} & 82.1 & \underline{96.3} & 98.0 & \textbf{65.9} & \textbf{88.4} & \textbf{93.2} & \textbf{523.9} \\ 
	\bottomrule[1.1pt]
	\end{tabular}
	\label{tab:all-results-updated}
\end{table*}

\begin{table}[!t]
	\footnotesize
	\centering
	\setlength{\tabcolsep}{3pt}
	\caption{\textbf{Results on \textsc{MSCOCO} 5K Test Set.} Best results are in \textbf{bold}, and the second-best scores are \underline{underlined}.}
	\begin{tabular}{l|ccc|ccc|c}
	\toprule[1.1pt]
	\multirow{3}[4]{*}{Methods} & \multicolumn{7}{c}{\textsc{MSCOCO} 5K Test Set} \\ 
	\cmidrule(lr){2-8}
	& \multicolumn{3}{c|}{Image-to-Text} & \multicolumn{3}{c|}{Text-to-Image} & \multirow{2}[2]{*}{R@Sum} \\ 
	\cmidrule(lr){2-4} \cmidrule(lr){5-7}
	& R@1 & R@5 & R@10 & R@1 & R@5 & R@10 & \\ 
	\midrule
	\multicolumn{8}{c}{\textit{Fast R-CNN + BiGRU}} \\ 
	\midrule
	SCAN$\ast$~\cite{SCAN} & 50.4 & 82.2 & 90.0 & 38.6 & 69.3 & 80.4 & 410.9 \\ 
	VSRN$\ast$~\cite{VSRN} & 53.0 & 81.1 & 89.4 & 40.5 & 70.6 & 81.1 & 415.7 \\ 
	VSE$\infty$~\cite{VSE-infty} & 56.6 & 83.6 & 91.4 & 39.3 & 69.9 & 81.1 & 421.9 \\ 
	NAAF~\cite{NAAF} & 58.9 & 85.2 & 92.0 & 42.5 & 70.9 & 81.4 & 430.9 \\ 
	\rowcolor{gray!20}
	CMSF (Ours) & 58.5	& 85.3	& 92.5	& 42.5	& 72.0	& 82.2	& 432.0 \\
	\midrule
	\multicolumn{8}{c}{\textit{Faster R-CNN + BERT}} \\ 
	\midrule
	MMCA~\cite{MMCA} & 54.0 & 82.5 & 90.7 & 38.7 & 69.7 & 80.8 & 416.4 \\ 
	TERAN$\ast$~\cite{TERAN} & 59.3 & 85.8 & 92.4 & 45.1 & \textbf{76.4} & \underline{84.4} & 443.4 \\ 
	VSE$\infty$~\cite{VSE-infty} & 58.3 & 85.3 & 92.3 & 42.4 & 72.7 & 83.2 & 434.2 \\ 
	VSRN++$\ast$~\cite{VSRN++} & 54.7 & 82.9 & 90.9 & 42.0 & 72.2 & 82.7 & 425.4 \\ 
	CHAN~\cite{CHAN} & 59.8 & 87.2 & 93.3 & \underline{44.9} & 74.5 & 84.2 & 443.9 \\ 
	HREM~\cite{HREM} & 62.3 & \underline{87.6} & \underline{93.4} & 43.9 & 73.6 & 83.3 & 444.1 \\ 
	USER~\cite{USER} & \textbf{63.7} & 87.4 & \textbf{93.5} & 44.8 & 73.4 & 82.7 & 445.5 \\ 
	MaxMatch~\cite{MaxMatch} & \underline{63.3} & \textbf{87.9} & 93.2 & 44.2 & 73.9 & 83.9 & \textbf{446.5} \\ 
	\rowcolor{gray!20}
	CMSF (Ours) & 61.5 & 86.7 & 92.8 & \textbf{45.1} & \underline{75.0} & \textbf{84.6} & \underline{445.8} \\ 
	\bottomrule[1.1pt]
	\end{tabular}
	\label{tab:mscoco5k}
\end{table}

\subsection{Comparisons with State-of-the-Art Methods}

\paragraph{Quantitative Comparison.} 
We compare CMSF with recent state-of-the-art ANN-based methods on both benchmarks, as shown in \cref{tab:all-results-updated,tab:mscoco5k}. Unlike SCAN~\cite{SCAN}, VSRN~\cite{VSRN}, TERAN~\cite{TERAN}, and VSRN++~\cite{VSRN++}, which boost performance via model ensembling, CMSF reports single-model results, similar to CHAN. Methods are grouped by replaceable feature-extraction backbones, where stronger extractors can lead to better performance (\textit{e.g.}, BERT and BiGRU). Our CMSF uniquely adopts a spike-driven framework. CMSF attains R@Sum scores of 533.8 and 523.9 on the \textsc{MSCOCO} 1K and \textsc{Flickr30K} 1K test sets, respectively, outperforming all ANN baselines. On the more challenging \textsc{MSCOCO} 5K test set (see \cref{tab:mscoco5k}), CMSF remains highly competitive and surpasses several state-of-the-art methods on multiple metrics. Compared with CHAN, CMSF improves every R@$K$ metric in text-to-image retrieval, confirming the benefit of Bidirectional Hard Alignment. Relative to attention-based approaches such as SCAN and MMCA, CMSF achieves 12.6\% and 7\% higher R@Sum on \textsc{Flickr30K}, highlighting the effectiveness of spike self-attention and cross-modal spike fusion. Although MaxMatch slightly outperforms CMSF in R@Sum on the larger \textsc{MSCOCO} 5K set, its use of the Hungarian algorithm for minimum-cost matching incurs significantly higher computational complexity. Leveraging the ability of SNNs to capture temporal dependencies between words, CMSF achieves particularly strong text-to-image retrieval performance.

\begin{figure}[!t]
	\centering
	\includegraphics[width = \linewidth]{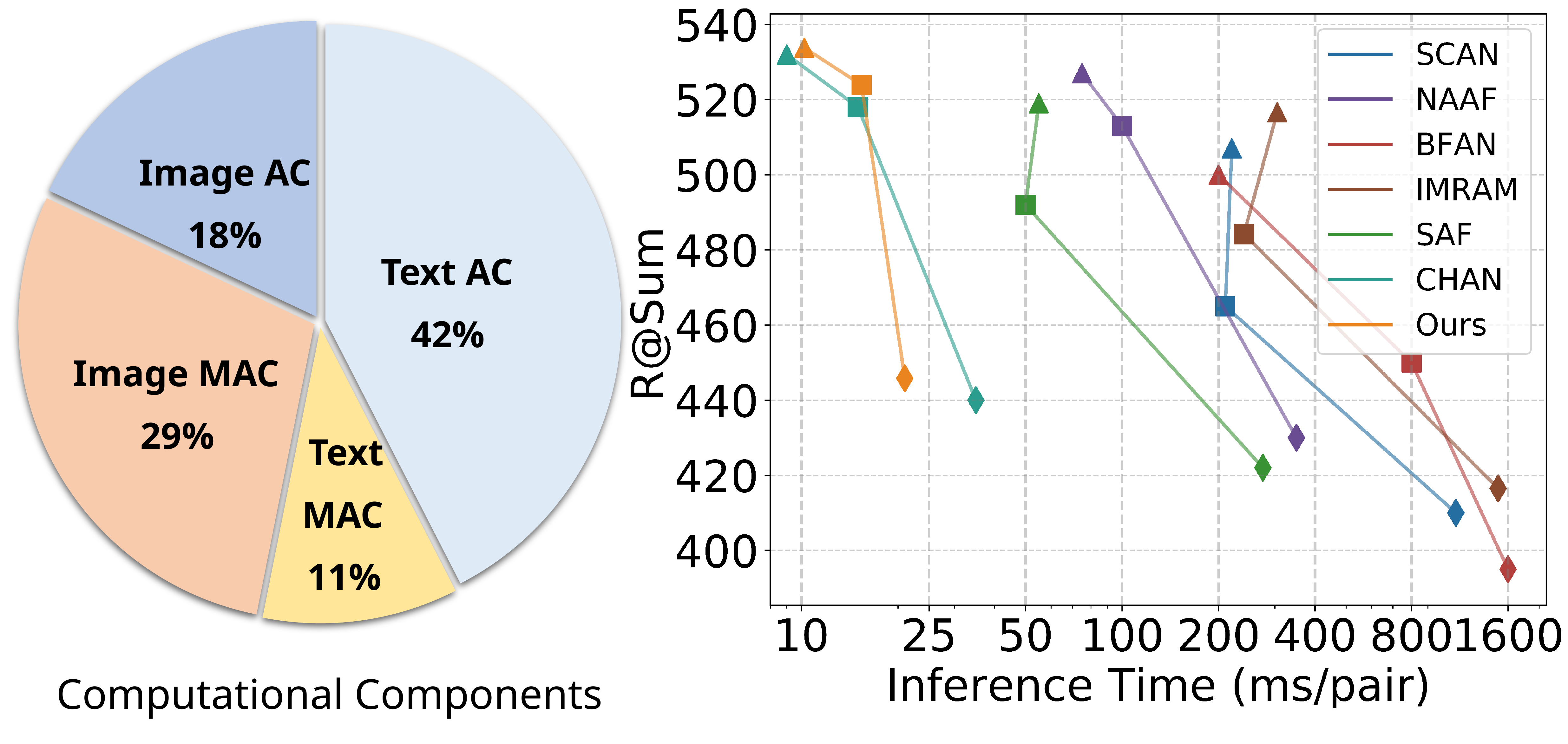} 
	\caption{\textbf{Computation profile of CMSF’s encoding stage (left) and inference speed comparison with ANN models (right).} AC: accumulate, MAC: multiply-and-accumulate. $\triangle$, $\square$, and $\lozenge$ represent results on \textsc{MSCOCO} 1K, \textsc{Flickr30K}, and \textsc{MSCOCO} 5K, respectively. The vertical axis shows R@Sum and the horizontal axis shows retrieval time (closer to the upper-left corner is better).}
	\label{fig:infer-time}
\end{figure}

\vspace{-8pt}
\paragraph{Inference Speed Comparison.}
CMSF employs a single SSA block, a minimal temporal step of $T = 2$, and lightweight pooling operations, and omits the Spike Fusion module during inference, thereby achieving high efficiency. As shown in \cref{fig:infer-time}(right), CMSF attains the fastest retrieval speed on \textsc{MSCOCO} 1K and processes each pair in just 21\,ms on \textsc{MSCOCO} 5K, outperforming CHAN~\cite{CHAN}. All results are measured on identical GPU hardware, where binary spike operations are still executed in floating-point form. As illustrated in \cref{fig:infer-time}(left), approximately 60\% of computations in the network are AC operations, and 40\% are MAC operations. True event computations on neuromorphic chips would further accelerate inference~\cite{SeqSNN}. Moreover, the mask operations in CMSF introduce negligible overhead, further improving runtime efficiency.

\vspace{-8pt}
\paragraph{Theoretical Energy Consumption Comparison.}
A key advantage of SNNs is their low energy consumption at inference. Following~\cite{Zhou_spikformer,LV-SpikeBERT}, we estimate theoretical energy usage of SNNs and ANNs on a 45\,nm neural chip~\cite{chip-45nm}, keeping all hyperparameters identical. Detailed computation formulas are provided in the \underline{Supplementary Material~C}.

\begin{table}[!t]
	\centering
	\footnotesize
	\setlength{\tabcolsep}{2pt}
	\caption{\textbf{Theoretical inference energy comparison.} Ops denote synaptic operations for SNNs and floating-point operations for ANNs. Best results are in bold.}
	\begin{tabular}{l|ccccc}
	\toprule[1.1pt]
	Methods & Architecture & {Param (M)} & {Ops (M)} & {Energy (mJ)} & R@Sum \\ 
	\midrule
	VSE$\infty$~\cite{VSE-infty} & RNN & \textbf{113.97} & \textbf{162.87} & 0.749 & 513.5 \\ 
	USER~\cite{USER} & RNN & 122.37 & 318.05 & 1.463 & 519.9 \\ 
	\midrule
	HREM~\cite{HREM} & Vanilla-Attn & 126.55 & 621.50 & 2.859 & 520.4 \\ 
	\rowcolor{gray!20}
	CMSF (Ours) & Spike-Attn & 179.30 & 265.04 & \textbf{0.626} & \textbf{523.9} \\ 
	\bottomrule[1.1pt]
	\end{tabular}
	\label{tab:energy}
\end{table}

For a fair comparison, we evaluate only the core retrieval networks, excluding the decoupled feature extractors, against state-of-the-art ANN models in \cref{tab:energy}. Compared with the RNN-based VSE$\infty$~\cite{VSE-infty}, both HREM~\cite{HREM} and CMSF introduce additional operations and parameters due to attention architectures. However, CMSF’s sparse spike activations reduce operations by 42\% relative to HREM, yielding the lowest theoretical energy of 0.626\,mJ, 78\% lower than HREM, while still achieving the highest recall.

\begin{table}[!t]
	\centering
	\footnotesize
	\setlength{\tabcolsep}{3pt}
	\caption{\textbf{Effect of alignment strategies on \textsc{Flickr30K} 1K test set.} Best results are in bold.}
	\begin{tabular}{l|ccc|ccc|c}
	\toprule[1.1pt]
	\multirow{2}[2]{*}{\makecell{Alignment \\ Method}} & \multicolumn{3}{c|}{Image-to-Text} & \multicolumn{3}{c|}{Text-to-Image} & \multirow{2}[2]{*}{R@Sum} \\ 
	\cmidrule(lr){2-4} \cmidrule(lr){5-7}
	& R@1 & R@5 & R@10 & R@1 & R@5 & R@10 & \\ 
	\midrule
	LSE & 75.9 & 93.6 & 96.7 & 59.5 & 84.5 & 90.4 & 500.6 \\ 
	VHA & 77.2 & 95.4 & 98.1 & 61.4 & 86.9 & 92.3 & 511.3 \\ 
	THA & 76.3 & 94.8 & 96.7 & 61.0 & 86.2 & 91.8 & 506.9 \\ 
	\rowcolor{gray!20}
	BiHA (Ours) & \textbf{82.1} & \textbf{96.3} & \textbf{98.0} & \textbf{65.9} & \textbf{88.4} & \textbf{93.2} & \textbf{523.9} \\ 
	\bottomrule[1.1pt]
	\end{tabular}
	\label{tab:Bi-direction}
\end{table}

\subsection{Ablation Studies}
\label{subsec:4.3}

\paragraph{Effect of Bidirectional Hard Alignment.}
We evaluate different alignment strategies on \textsc{Flickr30K} (see \cref{tab:Bi-direction}). Direct LogSumExp (LSE) pooling over the fine-grained similarity matrix yields an R@Sum of 500.6. Following CHAN~\cite{CHAN}, visual hard alignment (VHA), which selects the most relevant word for each region by taking the maximal similarity along the visual dimension, improves R@Sum to 511.3. Conversely, textual hard alignment (THA), which selects the most relevant region for each word, achieves 506.9. Our Bidirectional Hard Alignment (BiHA) integrates maximal similarities from both directions and achieves the best R@Sum score of 523.9.

\begin{table}[!t]
	\footnotesize
	\centering
	\setlength{\tabcolsep}{5pt}
	\caption{\textbf{Ablation of Early Alignment and Spike Fusion Soft Alignment on \textsc{Flickr30K} 1K Test Set.} Best results are in bold.}
	\begin{tabular}{cc|ccc}
	\toprule[1.1pt]
	Early-Align & Spike Fusion & R@1 (I2T) & R@1 (T2I) & R@Sum \\
	\midrule
	\Circle & \Circle & 76.5 & 63.0 & 511.6 \\
	\Circle & \CIRCLE & 78.5 & 63.3 & 514.2 \\
	\CIRCLE & \Circle & 79.7 & 64.1 & 517.0 \\
	\rowcolor{gray!20}
	\CIRCLE & \CIRCLE & \textbf{82.1} & \textbf{65.9} & \textbf{523.9} \\
	\bottomrule[1.1pt]
	\end{tabular}
	\label{tab:fusion}
\end{table}

\vspace{-8pt}
\paragraph{Effect of Early Alignment and Spike Fusion.}
To assess the contributions of early alignment and Spike Fusion Soft-label Alignment, we perform ablation experiments on \textsc{Flickr30K} (see \cref{tab:fusion}) with four variants: 
(1) Dual-stream only (no early alignment or fusion), R@Sum = 511.6; 
(2) Dual-stream + fusion only, R@Sum = 514.2; 
(3) Dual-stream + early alignment only, R@Sum = 517.0; 
(4) Full CMSF (dual-stream + early alignment + Spike Fusion), R@Sum = 523.9. 
When both modules are enabled, as illustrated in \cref{fig:motivation}, CMSF forms a complete brain-inspired multimodal SNN in which unimodal encoders benefit from joint upstream (early) and downstream (soft-label) optimization, achieving the best performance. This demonstrates that Spike Fusion provides multimodal, semantically enriched soft labels that effectively compensate for information lost during unimodal spike encoding by exploiting inter-modal interactions.

\begin{figure}[!t]
	\centering
	\includegraphics[width = \linewidth]{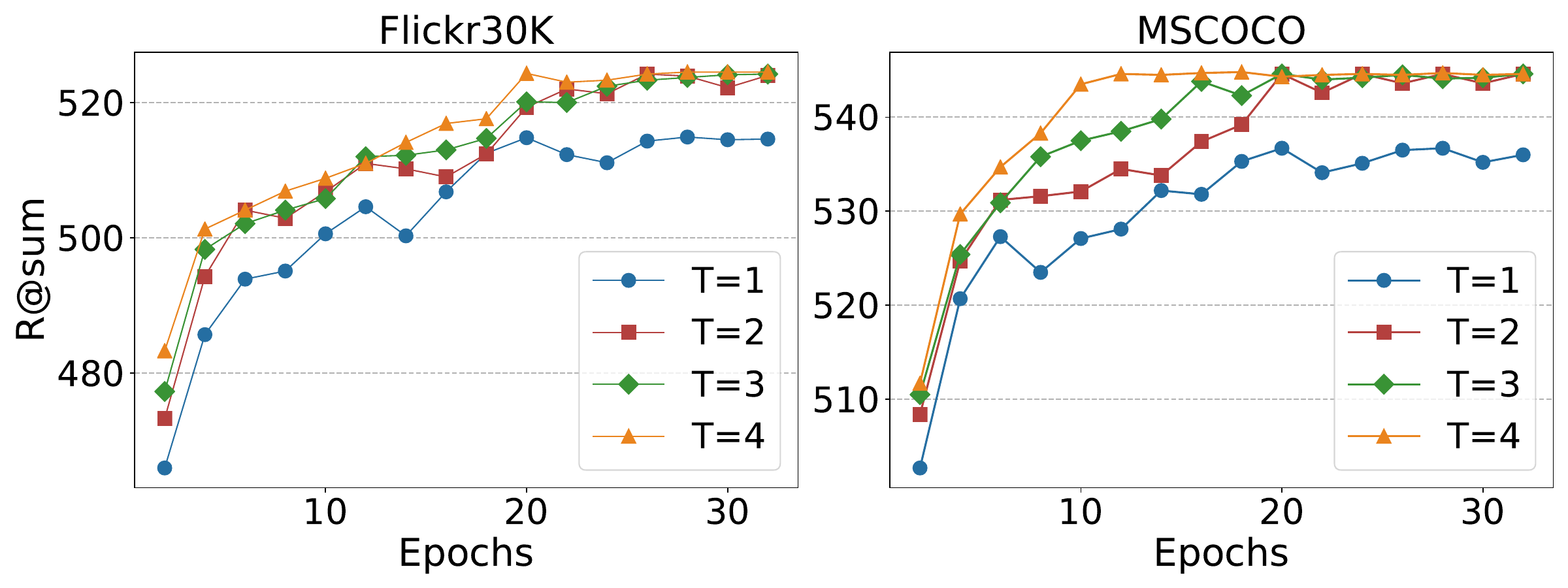} 
	\caption{\textbf{Effect of Different Time Steps.} CMSF performance across various time steps on \textsc{Flickr30K} and \textsc{MSCOCO} during training.}
	\label{fig:time-step}
\end{figure}

\vspace{-8pt}
\paragraph{Effect of Time Steps.}
In SNNs, larger time steps typically allow richer temporal dynamics. We investigate this in CMSF by varying $T \in \{1,2,3,4\}$ and analyzing the training process. As shown in \cref{fig:time-step}, when $T = 1$, spikes cannot be effectively retained or propagated, causing information loss, reduced accuracy, and unstable convergence due to higher sensitivity to noise. Increasing $T$ beyond 2 stabilizes training, accelerates convergence, and reduces fluctuations. However, more time steps introduce higher latency and energy consumption. Balancing accuracy, efficiency, and energy, we adopt $T = 2$ as the optimal configuration, achieving strong retrieval performance with low energy consumption and fast inference.

Additional ablation studies, parameter analyses, computational formulas, module variants, and qualitative retrieval visualizations are provided in the Supplementary Material.

\section{Conclusion}

In this work, we present a brain-inspired multimodal SNN framework for image-text retrieval (ITR). Extensive experiments show that the soft labels produced by our Spike Fusion mechanism effectively mitigate semantic information loss in SNNs, while the proposed Bidirectional Hard Alignment substantially enhances fine-grained cross-modal matching between visual and textual representations. CMSF achieves superior retrieval accuracy compared with state-of-the-art ANN-based methods, and, thanks to the intrinsic sparsity of spikes and the lightweight architecture, exhibits notably low energy consumption and high inference efficiency. Overall, this study broadens the applicability of SNNs to multimodal understanding and provides new insights into energy-efficient, biologically inspired cross-modal learning.


\section*{Acknowledgment}
This work was supported in part by the National Natural Science Foundation of China under Grants 62271361 and 62506011, the Hubei Provincial Key Research and Development Program under Grant 2024BAB039, and the China Postdoctoral Science Foundation under Grant GZB20250388.

{
 \small
 \bibliographystyle{ieeenat_fullname}
 \bibliography{main}

@String(AAAI = {AAAI})

@article{third-network-LIF,
  author  = {Wolfgang Maass},
  title   = {Networks of spiking neurons: The third generation of neural network models},
  journal = {Neural Networks},
  volume  = {10},
  number  = {9},
  pages   = {1659--1671},
  year    = {1997}
}

@inproceedings{SCAN,
  author    = {Kuang-Huei Lee and Xi Chen and Gang Hua and Houdong Hu and Xiaodong He},
  title     = {Stacked Cross Attention for Image-Text Matching},
  booktitle = {Proc. Eur. Conf. Comput. Vis.},
  pages     = {212--228},
  year      = {2018}
}

@inproceedings{VSE-infty,
  author    = {Jiacheng Chen and Hexiang Hu and Hao Wu and Yuning Jiang and Changhu Wang},
  title     = {Learning the Best Pooling Strategy for Visual Semantic Embedding},
  booktitle = {Proc. IEEE/CVF Conf. Comput. Vis. Pattern Recog.},
  pages     = {15789--15798},
  year      = {2021}
}

@inproceedings{VSE++,
  author    = {Fartash Faghri and David J. Fleet and Jamie Ryan Kiros and Sanja Fidler},
  title     = {VSE++: Improving Visual-Semantic Embeddings with Hard Negatives},
  booktitle = {Proc. Brit. Mach. Vis. Conf.},
  pages     = {12},
  year      = {2018}
}

@inproceedings{NAAF,
  author    = {Kun Zhang and Zhendong Mao and Quan Wang and Yongdong Zhang},
  title     = {Negative-Aware Attention Framework for Image-Text Matching},
  booktitle = {Proc. IEEE/CVF Conf. Comput. Vis. Pattern Recog.},
  pages     = {15640--15649},
  year      = {2022}
}

@inproceedings{VSRN,
  author    = {Kunpeng Li and Yulun Zhang and Kai Li and Yuanyuan Li and Yun Fu},
  title     = {Visual Semantic Reasoning for Image-Text Matching},
  booktitle = {Proc. IEEE/CVF Int. Conf. Comput. Vis.},
  pages     = {4653--4661},
  year      = {2019}
}

@inproceedings{bottom-up,
  author    = {Peter Anderson and Xiaodong He and Chris Buehler and Damien Teney and Mark Johnson and Stephen Gould and Lei Zhang},
  title     = {Bottom-Up and Top-Down Attention for Image Captioning and Visual Question Answering},
  booktitle = {Proc. IEEE/CVF Conf. Comput. Vis. Pattern Recog.},
  pages     = {6077--6086},
  year      = {2018}
}

@inproceedings{CLIP,
  author    = {Alec Radford and Jong Wook Kim and Chris Hallacy and Aditya Ramesh and Gabriel Goh and Sandhini Agarwal and Girish Sastry and Amanda Askell and Pamela Mishkin and Jack Clark and Gretchen Krueger and Ilya Sutskever},
  title     = {Learning Transferable Visual Models From Natural Language Supervision},
  booktitle = {Proc. Int. Conf. Mach. Learn.},
  pages     = {8748--8763},
  year      = {2021}
}

@article{Flickr30K,
  author  = {Peter Young and Alice Lai and Micah Hodosh and Julia Hockenmaier},
  title   = {From image descriptions to visual denotations: New similarity metrics for semantic inference over event descriptions},
  journal = {Trans. Assoc. Comput. Linguistics},
  volume  = {2},
  pages   = {67--78},
  year    = {2014}
}

@inproceedings{MSCOCO,
  author    = {Tsung{-}Yi Lin and Michael Maire and Serge J. Belongie and James Hays and Pietro Perona and Deva Ramanan and Piotr Doll{\'{a}}r and C. Lawrence Zitnick},
  title     = {Microsoft COCO: Common Objects in Context},
  booktitle = {Proc. Eur. Conf. Comput. Vis.},
  pages     = {740--755},
  year      = {2014}
}

@inproceedings{MoCo,
  author    = {Kaiming He and Haoqi Fan and Yuxin Wu and Saining Xie and Ross B. Girshick},
  title     = {Momentum Contrast for Unsupervised Visual Representation Learning},
  booktitle = {Proc. IEEE/CVF Conf. Comput. Vis. Pattern Recog.},
  pages     = {9726--9735},
  year      = {2020}
}

@inproceedings{Attention-is-all-you-need,
  author    = {Ashish Vaswani and Noam Shazeer and Niki Parmar and Jakob Uszkoreit and Llion Jones and Aidan N. Gomez and Lukasz Kaiser and Illia Polosukhin},
  title     = {Attention is All you Need},
  booktitle = {Adv. Neural Inform. Process. Syst.},
  pages     = {5998--6008},
  year      = {2017}
}

@article{Spikingjelly,
  author  = {Yunzhe Hao and Xuhui Huang and Meng Dong and Bo Xu},
  title   = {A biologically plausible supervised learning method for spiking neural networks using the symmetric STDP rule},
  journal = {Neural Networks},
  volume  = {121},
  pages   = {387--395},
  year    = {2020}
}

@article{Cao_CNN2SNN,
  author  = {Yongqiang Cao and Yang Chen and Deepak Khosla},
  title   = {Spiking Deep Convolutional Neural Networks for Energy-Efficient Object Recognition},
  journal = {Int. J. Comput. Vis.},
  volume  = {113},
  number  = {1},
  pages   = {54--66},
  year    = {2015}
}

@article{wu-STBP,
  author  = {Wu, Yujie and Deng, Lei and Li, Guoqi and Zhu, Jun and Shi, Luping},
  title   = {Spatio-temporal backpropagation for training high-performance spiking neural networks},
  journal = {Front. Neurosci.},
  volume  = {12},
  pages   = {331},
  year    = {2018}
}

@inproceedings{zheng-tdBN,
  author    = {Hanle Zheng and Yujie Wu and Lei Deng and Yifan Hu and Guoqi Li},
  title     = {Going Deeper With Directly-Trained Larger Spiking Neural Networks},
  booktitle = {Proc. AAAI Conf. Artif. Intell.},
  pages     = {11062--11070},
  year      = {2021}
}

@article{sengupta-res-snn,
  author  = {Abhronil Sengupta and Yuting Ye and Robert Wang and Chiao Liu and Kaushik Roy},
  title   = {Going deeper in spiking neural networks: VGG and residual architectures},
  journal = {Front. Neurosci.},
  volume  = {13},
  pages   = {95},
  year    = {2019}
}

@inproceedings{Zhou_spikformer,
  author    = {Zhaokun Zhou and Yuesheng Zhu and Chao He and Yaowei Wang and Shuicheng Yan and Yonghong Tian and Li Yuan},
  title     = {Spikformer: When Spiking Neural Network Meets Transformer},
  booktitle = {Proc. Int. Conf. Learn. Represent.},
  year      = {2023}
}

@inproceedings{QKformer,
  author    = {Chenlin Zhou and Han Zhang and Zhaokun Zhou and Liutao Yu and Liwei Huang and Xiaopeng Fan and Li Yuan and Zhengyu Ma and Huihui Zhou and Yonghong Tian},
  title     = {QKFormer: Hierarchical Spiking Transformer using Q-K Attention},
  booktitle = {Adv. Neural Inform. Process. Syst.},
  year      = {2024}
}

@inproceedings{Spike-Driven-transformer-V2,
  author    = {Man Yao and Jiakui Hu and Tianxiang Hu and Yifan Xu and Zhaokun Zhou and Yonghong Tian and Bo Xu and Guoqi Li},
  title     = {Spike-driven Transformer V2: Meta Spiking Neural Network Architecture Inspiring the Design of Next-generation Neuromorphic Chips},
  booktitle = {Proc. Int. Conf. Learn. Represent.},
  year      = {2024}
}

@inproceedings{LV-SpikeCNN-test,
  author    = {Changze Lv and Jianhan Xu and Xiaoqing Zheng},
  title     = {Spiking Convolutional Neural Networks for Text Classification},
  booktitle = {Proc. Int. Conf. Learn. Represent.},
  year      = {2023}
}

@article{LV-SpikeBERT,
  author  = {Changze Lv and Tianlong Li and Jianhan Xu and Chenxi Gu and Zixuan Ling and Cenyuan Zhang and Xiaoqing Zheng and Xuanjing Huang},
  title   = {SpikeBERT: A Language Spikformer Learned from BERT with Knowledge Distillation},
  journal = {arXiv preprint arXiv:2308.15122},
  year    = {2023}
}

@article{USER,
  author  = {Yan Zhang and Zhong Ji and Di Wang and Yanwei Pang and Xuelong Li},
  title   = {USER: Unified Semantic Enhancement With Momentum Contrast for Image-Text Retrieval},
  journal = {IEEE Trans. Image Process.},
  volume  = {33},
  pages   = {595--609},
  year    = {2024}
}

@inproceedings{CUSA,
  author    = {Hailang Huang and Zhijie Nie and Ziqiao Wang and Ziyu Shang},
  title     = {Cross-Modal and Uni-Modal Soft-Label Alignment for Image-Text Retrieval},
  booktitle = {Proc. AAAI Conf. Artif. Intell.},
  pages     = {18298--18306},
  year      = {2024}
}

@inproceedings{MMCA,
  author    = {Xi Wei and Tianzhu Zhang and Yan Li and Yongdong Zhang and Feng Wu},
  title     = {Multi-Modality Cross Attention Network for Image and Sentence Matching},
  booktitle = {Proc. IEEE/CVF Conf. Comput. Vis. Pattern Recog.},
  pages     = {10938--10947},
  year      = {2020}
}

@inproceedings{SCAN-inspire1,
  author    = {Zhong Ji and Kexin Chen and Haoran Wang},
  title     = {Step-Wise Hierarchical Alignment Network for Image-Text Matching},
  booktitle = {Proc. Int. Joint Conf. Artif. Intell.},
  pages     = {765--771},
  year      = {2021}
}

@article{SCAN-inspire2,
  author  = {Yaodong Wang and Zhong Ji and Kexin Chen and Yanwei Pang and Zhongfei Zhang},
  title   = {COREN: Multi-Modal Co-Occurrence Transformer Reasoning Network for Image-Text Retrieval},
  journal = {Neural Process. Lett.},
  volume  = {55},
  number  = {5},
  pages   = {5959--5978},
  year    = {2023}
}

@article{SCAN-inspire3,
  author  = {Keyu Wen and Xiaodong Gu and Qingrong Cheng},
  title   = {Learning Dual Semantic Relations With Graph Attention for Image-Text Matching},
  journal = {IEEE Trans. Circuits Syst. Video Technol.},
  volume  = {31},
  number  = {7},
  pages   = {2866--2879},
  year    = {2021}
}

@inproceedings{SCO,
  author    = {Yan Huang and Qi Wu and Chunfeng Song and Liang Wang},
  title     = {Learning Semantic Concepts and Order for Image and Sentence Matching},
  booktitle = {Proc. IEEE/CVF Conf. Comput. Vis. Pattern Recog.},
  pages     = {6163--6171},
  year      = {2018}
}

@inproceedings{CAMP,
  author    = {Ying Zhang and Huchuan Lu},
  title     = {Deep Cross-Modal Projection Learning for Image-Text Matching},
  booktitle = {Proc. Eur. Conf. Comput. Vis.},
  pages     = {707--723},
  year      = {2018}
}

@inproceedings{GCN1,
  author    = {Chunxiao Liu and Zhendong Mao and Tianzhu Zhang and Hongtao Xie and Bin Wang and Yongdong Zhang},
  title     = {Graph Structured Network for Image-Text Matching},
  booktitle = {Proc. IEEE/CVF Conf. Comput. Vis. Pattern Recog.},
  pages     = {10918--10927},
  year      = {2020}
}

@article{GCN2,
  author  = {Ya Jing and Wei Wang and Liang Wang and Tieniu Tan},
  title   = {Learning Aligned Image-Text Representations Using Graph Attentive Relational Network},
  journal = {IEEE Trans. Image Process.},
  volume  = {30},
  pages   = {1840--1852},
  year    = {2021}
}

@article{VSRN++,
  author  = {Kunpeng Li and Yulun Zhang and Kai Li and Yuanyuan Li and Yun Fu},
  title   = {Image-Text Embedding Learning via Visual and Textual Semantic Reasoning},
  journal = {IEEE Trans. Pattern Anal. Mach. Intell.},
  volume  = {45},
  number  = {1},
  pages   = {641--656},
  year    = {2023}
}

@article{FasterRCNN,
  author  = {Shaoqing Ren and Kaiming He and Ross B. Girshick and Jian Sun},
  title   = {Faster R-CNN: Towards Real-Time Object Detection with Region Proposal Networks},
  journal = {IEEE Trans. Pattern Anal. Mach. Intell.},
  volume  = {39},
  number  = {6},
  pages   = {1137--1149},
  year    = {2017}
}

@inproceedings{SeqSNN,
  author    = {Changze Lv and Yansen Wang and Dongqi Han and Xiaoqing Zheng and Xuanjing Huang and Dongsheng Li},
  title     = {Efficient and Effective Time-Series Forecasting with Spiking Neural Networks},
  booktitle = {Proc. Int. Conf. Mach. Learn.},
  year      = {2024}
}

@inproceedings{chip-45nm,
  author    = {Mark Horowitz},
  title     = {1.1 Computing's Energy Problem (and What We Can Do About It)},
  booktitle = {Proc. IEEE Int. Conf. Solid-State Circuits},
  pages     = {10--14},
  year      = {2014}
}

@inproceedings{BERT,
  author    = {Jacob Devlin and Ming-Wei Chang and Kenton Lee and Kristina Toutanova},
  title     = {BERT: Pre-training of Deep Bidirectional Transformers for Language Understanding},
  booktitle = {Proc. Conf. North Am. Chapter Assoc. Comput. Linguist.},
  pages     = {4171--4186},
  year      = {2019}
}

@inproceedings{ONESTEP,
  author    = {Xiaotian Song and Andy Song and Rong Xiao and Yanan Sun},
  title     = {One-step Spiking Transformer with a Linear Complexity},
  booktitle = {Proc. Int. Joint Conf. Artif. Intell.},
  pages     = {3142--3150},
  year      = {2024}
}

@inproceedings{CHAN,
  author    = {Zhengxin Pan and Fangyu Wu and Bailing Zhang},
  title     = {Fine-grained Image-text Matching by Cross-modal Hard Aligning Network},
  booktitle = {Proc. IEEE/CVF Conf. Comput. Vis. Pattern Recog.},
  pages     = {19275--19284},
  year      = {2023}
}

@inproceedings{SoftCLIP,
  author    = {Yuting Gao and Jinfeng Liu and Zihan Xu and Tong Wu and Enwei Zhang and Ke Li and Jie Yang and Wei Liu and Xing Sun},
  title     = {SoftCLIP: Softer Cross-Modal Alignment Makes CLIP Stronger},
  booktitle = {Proc. AAAI Conf. Artif. Intell.},
  pages     = {1860--1868},
  year      = {2024}
}

@article{Sensor,
  author  = {Jon Driver and Toemme Noesselt},
  title   = {Multisensory Interplay Reveals Crossmodal Influences on ‘Sensory-Specific’ Brain Regions, Neural Responses, and Judgments},
  journal = {Neuron},
  volume  = {57},
  number  = {1},
  pages   = {11--23},
  year    = {2008}
}

@inproceedings{HREM,
  author    = {Zheren Fu and Zhendong Mao and Yan Song and Yongdong Zhang},
  title     = {Learning Semantic Relationship among Instances for Image-Text Matching},
  booktitle = {Proc. IEEE/CVF Conf. Comput. Vis. Pattern Recog.},
  pages     = {15159--15168},
  year      = {2023}
}

@article{TERAN,
  author  = {Nicola Messina and Giuseppe Amato and Andrea Esuli and Fabrizio Falchi and Claudio Gennaro and Stéphane Marchand-Maillet},
  title   = {Fine-Grained Visual Textual Alignment for Cross-Modal Retrieval Using Transformer Encoders},
  journal = {ACM Trans. Multimedia Comput. Commun. Appl.},
  volume  = {17},
  number  = {4},
  pages   = {128:1--128:23},
  year    = {2021}
}

@inproceedings{MaxMatch,
  author    = {Hani Alomari and Anushka Sivakumar and Andrew Zhang and Chris Thomas},
  title     = {Maximal Matching Matters: Preventing Representation Collapse for Robust Cross-Modal Retrieval},
  booktitle = {Proc. Annu. Meet. Assoc. Comput. Linguist.},
  pages     = {31769--31785},
  year      = {2025}
}

@article{youhongANN2SNN,
  author  = {Hong You and Xian Zhong and Wenxuan Liu and Qi Wei and Wenxin Huang and Zhaofei Yu and Tiejun Huang},
  title   = {Converting Artificial Neural Networks to Ultralow-Latency Spiking Neural Networks for Action Recognition},
  journal = {IEEE Trans. Cogn. Dev. Syst.},
  volume  = {16},
  number  = {4},
  pages   = {1533--1545},
  year    = {2024}
}

@inproceedings{hushengwang,
  author    = {Xian Zhong and Shengwang Hu and Wenxuan Liu and Wenxin Huang and Jianhao Ding and Zhaofei Yu and Tiejun Huang},
  title     = {Towards Low-latency Event-based Visual Recognition with Hybrid Step-wise Distillation Spiking Neural Networks},
  booktitle = {Proc. ACM Int. Conf. Multimedia},
  pages     = {9828--9836},
  year      = {2024}
}

@inproceedings{liuwenxuan,
  author    = {Wenxuan Liu and Yao Deng and Kang Chen and Xian Zhong and Zhaofei Yu and Tie-Jun Huang},
  title     = {SOTA: Spike-Navigated Optimal Transport Saliency Region Detection in Composite-bias Videos},
  booktitle = {Proc. Int. Joint Conf. Artif. Intell.},
  year      = {2025}
}

@article{SUYOLO,
  author  = {Chenyang Li and Wenxuan Liu and Guoqiang Gong and Xiaobo Ding and Xian Zhong},
  title   = {SU-YOLO: Spiking Neural Network for Efficient Underwater Object Detection},
  journal = {Neurocomputing},
  volume  = {644},
  pages   = {130310},
  year    = {2025}
}

@inproceedings{GRNN,
  author    = {Junyoung Chung and {\c{C}}aglar G{\"{u}}l{\c{c}}ehre and Kyunghyun Cho and Yoshua Bengio},
  title     = {Gated Feedback Recurrent Neural Networks},
  booktitle = {Proc. Int. Conf. Mach. Learn.},
  pages     = {2067--2075},
  year      = {2015}
}

@inproceedings{gmlp,
  author    = {Hanxiao Liu and Zihang Dai and David R. So and Quoc V. Le},
  title     = {Pay Attention to MLPs},
  booktitle = {Adv. Neural Inform. Process. Syst.},
  pages     = {9204--9215},
  year      = {2021}
}

@article{excitation-inhibition,
  author  = {Liang, Junhao and Yang, Zhuda and Zhou, Changsong},
  title   = {Excitation--inhibition balance, neural criticality, and activities in neuronal circuits},
  journal = {Neuroscientist},
  volume  = {31},
  number  = {1},
  pages   = {31--46},
  year    = {2025}
}

@inproceedings{spike2former,
  title={Spike2former: Efficient spiking transformer for high-performance image segmentation},
  author={Lei, Zhenxin and Yao, Man and Hu, Jiakui and Luo, Xinhao and Lu, Yanye and Xu, Bo and Li, Guoqi},
  booktitle={Proc. AAAI Conf. Artif. Intell.},
  volume={39},
  number={2},
  pages={1364--1372},
  year={2025}
}

@inproceedings{spikeyolo,
  title={Integer-valued training and spike-driven inference spiking neural network for high-performance and energy-efficient object detection},
  author={Luo, Xinhao and Yao, Man and Chou, Yuhong and Xu, Bo and Li, Guoqi},
  booktitle={Proc. Eur. Conf. Comput. Vis.},
  pages={253--272},
  year={2024},
}
}

\clearpage
\setcounter{page}{1}
\maketitlesupplementary
\setcounter{section}{0}     

\renewcommand\thesection{\Alph{section}}
\section{Spike Fusion Methods}
\label{sec:appendixA}
To validate the effectiveness of our Spike Fusion, we design multiple fusion strategies: Spike Cross Attention (SCA), Spike-Concat Self Attention (SCSA), and Spike Comb Cross Attention (SCCA). In this section, we present detailed experiments and analysis of the SCA and SCSA methods.

\subsection{Spike Cross Attention}

Drawing inspiration from Spikformer~\cite{Zhou_spikformer}, we propose a purely spiking-driven network called Spike Cross Attention (SCA). Its structure mirrors the conventional cross-attention mechanism, but replaces ReLU activations with LIF neurons and omits the Softmax operation to better align with SNN characteristics.

\begin{figure}[!t]
	\centering
	\includegraphics[width = \linewidth]{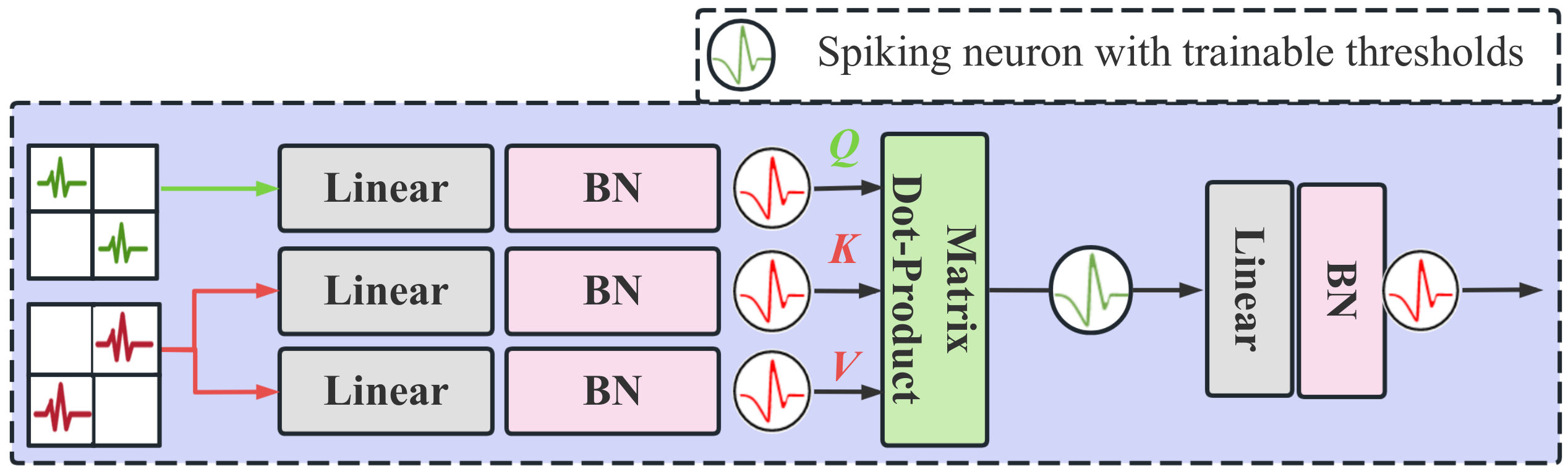} 
	\caption{\textbf{Detailed structure of our Spike Cross Attention.} Embeddings from both modalities are fused at the spike level via cross-attention matrix multiplication.}
	\label{fig:spike-cross-attention}
\end{figure}

As shown in \cref{fig:spike-cross-attention}, the image spike embedding (green) serves as the \textit{Query}, while the text spike embedding (red) provides both the \textit{Key} and \textit{Value}. All inputs firstly pass through a \{\texttt{Linear}, \texttt{BN}, \texttt{LIF}\} layer before SCA’s matrix operations. During this process, salient elements of the Key activate corresponding \textit{Query} spikes, joint spiking information is retained, and redundant \textit{Query} activations are suppressed. The resulting \textit{Query} spike matrix is thereby fused with cross-modal information while preserving sparsity.

SCA is specifically designed for cross-modal spike sequences: since $Q$, $K$, and $V$ are binary spike matrices, their ``dot products'' reduce to logical AND ($\&$) followed by summation, meeting SNN requirements. Moreover, the order of computation, $(Q K^\top )V$ \textit{vs.} $Q(K^\top V)$, can be chosen dynamically to minimize time complexity, selecting between $O(N^2 D)$ and $O(N D^2)$. Thus, SCA maintains both biological plausibility and computational efficiency throughout the spike fusion process.

\subsection{Spike-Concat Self Attention}

To assess the impact of single-stream versus dual-stream architectures on our Spike Fusion, we introduce Spike-Concat Self Attention (SCSA), a single-stream variant. SCSA’s structure parallels SCA, with the key difference being input handling: image and text spike embeddings are concatenated along the $N$ and $L$ dimensions before entering the SCSA block:
\begin{align}
	\bm{X}_{\mathrm{concat}} = \binom{R}{E} = \left\{r_1;\dots;r_N;e_1;\dots;e_L \right\},
\end{align}

where $\bm{X}_{\mathrm{concat}} \in \mathbb{R}^{T\times(N+L)\times D}$. Omitting Softmax and scaling factors, SCSA is defined as:
\begin{align}
	\mathrm{SCSA} \left(Q_S,K_S,V_S\right) = Q_S K_S^\top V_S,
\end{align}
where $Q_S$, $K_S$, and $V_S$ are obtained by linearly projecting $X_{\mathrm{concat}}$:
\begin{align}
	Q_S & = X_\mathrm{concat} W^Q = \binom{R W^Q}{E W^Q} = \binom{Q_R}{Q_E}, \\
	K_S & = X_\mathrm{concat} W^K = \binom{R W^K}{E W^K} = \binom{K_R}{K_E}, \\
	V_S & = X_\mathrm{concat} W^V = \binom{R W^V}{E W^V} = \binom{V_R}{V_E}.
\end{align}
By matrix multiplication rules:
\begin{align}
\begin{aligned}
	Q_S K_S^\top V_S & = \binom{Q_R}{Q_E} \left(K_R^\top K_E^\top \right) \cdot \binom{V_R}{V_E} \\
	& = \binom{Q_R K_R^\top Q_R K_E^\top }{Q_EK_R^\top Q_EK_E^\top } \cdot \binom{V_R}{V_E} \\
	& = \binom{Q_R K_R^\top V_R + Q_R K_E^\top V_E}{Q_E K_E^\top V_E + Q_E K_R^\top V_R}.
\end{aligned}
\end{align}
The final SCSA output $\tilde{X}\in\mathbb{R}^{T\times(N+L)\times D}$ is then split according to the original modality dimensions, yielding the fused spike embeddings: 
\begin{align}
	R = Q_R K_R^\top V_R + Q_R K_E^\top V_E \in \mathbb{R}^{T \times N \times D}, \\
	E = Q_E K_E^\top V_E + Q_E K_R^\top V_R \in \mathbb{R}^{T \times L \times D}.
\end{align}
This ensures that the outputs $R$ and $E$ both integrate cross-modal information and preserve their own modality-specific features, thereby supporting effective intra- and inter-modal alignment.

\begin{table}[!t]
	\centering
	\setlength{\tabcolsep}{12pt}
	\footnotesize
    \caption{\textbf{Time and space complexities of SCA, SCSA, and SCCA.} $N$ (regions) and $L$ (words) are of similar magnitude.}
	\begin{tabular}{l|ll}
	\toprule[1.1pt]
	Method & Time Complexity & Space Complexity \\
	\midrule
	SCA & $O(ND^2)$ & $O(D^2 + ND)$ \\
	SCSA & $O((N+L)D^2)$ & $O(D^2 + (N+L)D)$ \\
	SCCA & $O(N)$ & $O(D)$ \\
	\bottomrule[1.1pt]
	\end{tabular}
	\label{tab:time-space}
\end{table}

\subsection{Comparison Results
}
In \cref{tab:time-space}, we compare the time and space complexities of our three fusion methods. For SCA and SCSA, the matrix multiplication step incurs $O(ND^2)$ time complexity, and, because both the attention map and the Query matrix must be stored, $O(D^2 + ND)$ space complexity. By contrast, mask-based Spike Comb Cross Attention traverses only $N/h$ elements per comb and stores just $h$ combs, reducing time complexity to $O(N)$ and space complexity to $O(D)$.

\begin{table}[!t]
	\footnotesize
	\centering
	\setlength{\tabcolsep}{4pt}
    \caption{\textbf{Results of different fusion methods on \textsc{Flickr30K} 1K test set.} R@K denotes Recall@K, and R@Sum is the sum of R@1, R@5, and R@10 for both retrieval directions.}
	\begin{tabular}{l|ccc|ccc|c}
	\toprule[1.1pt]
	\multirow{2}[2]{*}{Methods} & \multicolumn{3}{c|}{Image-to-Text} & \multicolumn{3}{c|}{Text-to-Image} & \multirow{2}[2]{*}{R@Sum} \\
	\cmidrule(lr){2-4} \cmidrule(lr){5-7}
	& R@1 & R@5 & R@10 & R@1 & R@5 & R@10 & \\
	\midrule
	SCA & 81.3 & 95.8 & \textbf{98.5} & 64.9 & 88.3 & \textbf{93.4} & 522.1 \\
	SCSA & 81.0 & \textbf{96.4} & 98.1 & 64.9 & \textbf{88.4} & 93.2 & 522.0 \\
	\rowcolor{gray!20}
	SCCA & \textbf{82.1} & 96.3 & 98.0 & \textbf{65.9} & \textbf{88.4} & 93.2 & \textbf{523.9} \\
	\bottomrule[1.1pt]
	\end{tabular}
	
	\label{tab:3fusion}
\end{table}

In \cref{tab:3fusion}, we compare the performance of three Spike Fusion methods on \textsc{Flickr30K} under identical settings. The SCCA-based method achieves the highest accuracy, while the performance gap among the three approaches remains small, demonstrating the effectiveness of spike fusion for cross-modal interaction. Additionally, during training SCCA consumes less memory due to its simpler and more efficient network structure.

\section{Pre-trained VLMs}
\label{sec:appendixB}
As a general-purpose model, the pretrained Vision-Language Models treat image-text retrieval merely as one of their training objectives to enhance generalization capability. Pretrained models such as CLIP~\cite{CLIP}, which achieve impressive performance through large-scale data and deep network architectures, are not comparable in retrieval efficiency or energy consumption to specially designed lightweight retrieval models~\cite{USER}.

\begin{table}[!t]
	\footnotesize
	\centering
	\setlength{\tabcolsep}{2pt}
	\caption{\textbf{Results Compared to Pre-trained Methods.} frcnn means FasterRCNN~\cite{FasterRCNN}.}
	\begin{tabular}{l|c|ccc|ccc|c}
	\toprule[1.1pt]
	\multirow{2}[2]{*}{Methods} & \multirow{2}[2]{*}{Structure} 
	& \multicolumn{3}{c|}{Image-to-Text} & \multicolumn{3}{c|}{Text-to-Image} & \multirow{2}[2]{*}{R@Sum} \\ 
	\cmidrule(lr){3-5} \cmidrule(lr){6-8}
	& &R@1 & R@5 & R@10 & R@1 & R@5 & R@10 & \\ 
    \midrule
	\multicolumn{9}{c}{\textit{Flickr30K 1K Test Set}} \\ 
	\midrule
    CLIP & ViT-B/32&78.7 &95.4& 98.0& 66.3& 88.6& 93.1 &520.0 \\ 
    CLIP & ViT-L/14 &87.3& 99.0& 99.5& 76.4& 94.8 &97.4 &554.5\\
    \rowcolor{gray!20}
    CMSF & frcnn+bigru& 80.7 & 95.0	& 97.6	& 61.3	& 85.9	& 91.3	& 511.8\\
    \rowcolor{gray!20}
    CMSF & frcnn+bert& 82.1 & 96.3 & 98.0 & 65.9 & 88.4 & 93.2 & 523.9 \\

	\midrule
	\multicolumn{9}{c}{\textit{MSCOCO 5K Test Set}} \\ 
	\midrule
	CLIP & ViT-B/32&56.3& 81.7 &89.4& 42.8 &71.2 &81.1 &422.6\\
     CLIP & ViT-L/14 &67.1& 89.4 &94.7 &51.6 &79.1 &87.7& 469.6 \\
     \rowcolor{gray!20}
    CMSF & frcnn+bigru& 58.5	& 85.3	& 92.5	& 42.5	& 72.0	& 82.2	& 432.0  \\
    \rowcolor{gray!20}
	CMSF & frcnn+bert&61.5 & 86.7 & 92.8 & 45.1 & 75.0 & 84.6 & 445.8 \\ 
	\bottomrule[1.1pt]
	\end{tabular}
	\label{tab:clip}
\end{table}

While CLIP employs a ViT backbone to extract image and text features, our method and baselines use Faster R-CNN and BERT for feature extraction, respectively. To further validate the effectiveness of our approach, we also conducted comparison experiments with CLIP. As shown in \cref{tab:clip} , our method (with a single attention block) significantly outperforms CLIP’s base variant (12 Transformer blocks) but falls slightly behind the deeper large variant (24 Transformer blocks). This result also indicates that the feature extraction network is interchangeable, and a stronger backbone can lead to improved performance.

\section{Theoretical Energy Consumption}

The computational energy consumption on neuromorphic hardware is often measured by operation counts. In ANNs, each operation involves floating-point multiplications and additions (MACs), and the computational cost is estimated by floating-point operations (FLOPs). SNNs, however, are more energy-efficient on neuromorphic hardware since neurons perform only accumulation computations (AC) during spikes, counted as synaptic operations (SyOPs). Following~\cite{SeqSNN}, the theoretical energy consumption of SNN layer $l$ is:
\begin{align}
	\mathrm{Energy} \left(l \right) = E_{\mathrm{AC}} \times \mathrm{SOP}_s \left(l \right).
\end{align}
Analogously, for an ANN layer $f$, the theoretical energy consumption is:
\begin{align}
	\mathrm{Energy} \left(f \right) = E_{\mathrm{MAC}} \times \mathrm{FLOP}_s \left(f \right).
\end{align}
We assume MAC and AC operations on 45 nm hardware~\cite{chip-45nm}, with $E_{\mathrm{MAC}} = 4.6$ pJ and $E_{\mathrm{AC}} = 0.9$ pJ (1 J = 10$^3$ mJ = 10$^12$ pJ). The number of synaptic operations in SNN layer $l$ is estimated as
\begin{align}
	\mathrm{SOP}_s \left(l \right) = T \times \mathrm{Rate} \times \mathrm{FLOP}_s \left(l \right),
\end{align}
where $T$ is the number of time steps and $\mathrm{Rate}$ is the firing rate of the input spike train at layer $l$.

\begin{table}[!t]
	\centering
	\footnotesize 
	\setlength{\tabcolsep}{12pt}
	\caption{\textbf{Detailed calculation formulas.} The theoretical energy consumption of each CMSF layer.}
	\begin{tabular}{l|l}
	\toprule[1.1pt]
	Neural Layer & Theoretical Consumption \\
	\midrule
	Region Linear & $E_{\mathrm{AC}}\cdot T \cdot R_{r} \cdot FL_{r}$ \\
	Word Linear & $E_{\mathrm{AC}}\cdot T \cdot R_{w} \cdot FL_{w}$ \\
	\midrule
	Q, K, V & $E_{\mathrm{AC}}\cdot T \cdot R_{0} \cdot 3ND^2$ \\
	SCA & $E_{\mathrm{AC}}\cdot T_{1} \cdot R_{1} \cdot ND^2$ \\
	SCSA & $E_{\mathrm{AC}}\cdot T_{2} \cdot R_{2} \cdot (N+L)D^2$ \\
	SCCA & $E_{\mathrm{AC}}\cdot T_{3} \cdot R_{3} \cdot N$ \\
	Out Linear & $E_{\mathrm{AC}}\cdot T \cdot R_{o} \cdot FL_{o}$ \\
	\midrule
	Gate Linear & $E_{\mathrm{AC}}\cdot T \cdot R_{g} \cdot FL_{g}$ \\
	Gate Multiply & 0 \\
	\bottomrule[1.1pt]
	\end{tabular}
	
	\label{tab:theoretical}
\end{table}

In \cref{tab:theoretical}, we present the theoretical energy consumption formulas for each spiking neuron layer in our CMSF network. This includes the Linear layers used to align input dimensions, the Linear projections for Q, K, and V matrices, the matrix multiplications in the three Spike Fusion schemes (SCA, SCSA, SCCA), and the output projection Linear layers. Notably, for the Spike Gated MLP, the gated matrix multiplication acts as a masking operation and therefore incurs negligible energy cost on neuromorphic hardware.

\begin{table}[!t]
	\centering
	\footnotesize
	\setlength{\tabcolsep}{4pt}
    \caption{\textbf{Ablation of Spike Generator designs on Flickr30K 1K.} R@K denotes Recall@K for image-to-text (first three columns) and text-to-image (next three columns); R@Sum is the total.}
	\begin{tabular}{l|ccc|ccc|c}
	\toprule[1.0pt]
	\multirow{2}[2]{*}{\makecell{Spike \\ Generator}} & \multicolumn{3}{c|}{Image-to-Text} & \multicolumn{3}{c|}{Text-to-Image} & \multirow{2}[2]{*}{R@Sum} \\
	\cmidrule(lr){2-4} \cmidrule(lr){5-7}
	& R@1 & R@5 & R@10 & R@1 & R@5 & R@10 & \\
	\midrule
	Conv-BN & 77.9 & 95.8 & \textbf{98.6} & 60.8 & 85.8 & 91.5 & 510.4 \\
	Delta-BN & 76.1 & 95.4 & 97.8 & 58.5 & 84.0 & 90.2 & 502.0 \\
	Linear-BN & 80.3 & \textbf{96.5} & 98.2 & 64.8 & 87.5 & 92.7 & 520.0 \\
	Linear-LN & 81.3 & 96.1 & 97.7 & 64.4 & \textbf{88.6} & 93.2 & 521.3 \\
	Repeat-BN &80.2 & 95.9 & 97.5 & 64.4 & 88.2 & \textbf{93.4} & 519.6\\
	\rowcolor{gray!20}
	Repeat-LN & \textbf{82.1} & {96.3} & {98.0} & \textbf{65.9} & {88.4} & {93.2} & \textbf{523.9} \\
	\bottomrule[1.1pt]
	\end{tabular}
	
	\label{tab:spike-encoder-ablation}
\end{table}

\section{Spike Generator}

The Spike Generator, positioned at the front of the network, converts continuous-valued inputs $X_f$ into discrete spike trains $X_s$ and plays a crucial role in preserving semantic information. A simple approach is to repeat the original features $T$ times before neuronal activation. Prior work, such as~\cite{SeqSNN}, has proposed Delta and Convolution Generators for mapping floating-point time-series data to spike trains; however, these methods may not effectively retain intra-modal semantic structure in multi-modal image-text retrieval tasks. To address this, we explore various combinations of dimensional-expansion and normalization techniques to identify an generator that best preserves semantic information for downstream processing. Our final design is:
\begin{align}
	X_{s} = \mathcal{SN} \left(\mathrm{LN} \left(\mathrm{Repeat} \left(X_{f},T \right) \right) \right),
\end{align} 
where $\mathrm{Repeat}(X_{f},T)$ duplicates features $T$ times, LN denotes layer normalization, and $\mathcal{SN}$ is the LIF neuron. This sequence can be represented as \{\texttt{Linear-LN-LIF}\} and abbreviated as \{\texttt{Linear-LN}\}. Ablation studies in \cref{tab:spike-encoder-ablation} confirm the effectiveness of this Spike Generator design. The experimental results demonstrate that the proposed method achieves significant improvements in both Recall@1 and Recall@Sum metrics.

\section{Comb Teeth}

Inspired by~\cite{QKformer}, we introduce Spike Comb Cross Attention to enable spike-level fusion. We split the \textit{Query} matrix into $h$ sub-matrices, sum each to obtain $h$ vectors, and then apply spiking neuron activation. These vectors reveal distribution patterns across the embedding dimension $D$, analogous to the ``teeth'' of a comb: when they ``comb through'' the \textit{Key} matrix, they align the spike distributions of \textit{Query} and \textit{Key}, encouraging both modalities to fire at corresponding positions. This alignment enhances cross-modal integration.

\begin{table}[!t]
	\centering
	\footnotesize
	\setlength{\tabcolsep}{4pt}
    \caption{\textbf{Ablation of comb head numbers on \textsc{Flickr30K} 1K.} R@K denotes Recall@K for image-to-text (columns 2-4) and text-to-image (columns 5-7); R@Sum is the total across all six metrics.}
	\begin{tabular}{l|ccc|ccc|c}
	\toprule[1.1pt]
	\multirow{2}[2]{*}{\makecell{Head \\ Num}} & \multicolumn{3}{c|}{Image-to-Text} & \multicolumn{3}{c|}{Text-to-Image} & \multirow{2}[2]{*}{R@Sum} \\
	\cmidrule(lr){2-4} \cmidrule(lr){5-7}
	& R@1 & R@5 & R@10 & R@1 & R@5 & R@10 & \\
	\midrule
	$h = 2$ & 79.4 & \textbf{96.5} & 97.7 & 63.9 & 88.0 & 93.0 & 518.6 \\
	$h = 3$ & 80.5 & 96.1 & 98.0 & 64.4 & \textbf{88.5} & \textbf{93.3} & 520.8\\
	$h = 4$ & 80.2 & 96.3 & \textbf{98.2} & 64.9 & 88.4 & 93.2 & 521.2 \\
	\rowcolor{gray!20}
	$h = 6$ & \textbf{82.1} & 96.3 & 98.0 & \textbf{65.9} & 88.4 & 93.2 & \textbf{523.9} \\
	$h = 9$ & 80.2 & 95.9 & 98.0 & 64.3 & 88.3 & 92.9 & 519.5\\
	$h = 12$ & 80.4 & 96.5 & 98.1 & 64.5 & 88.3 & 93.2 & 520.9 \\
	$h = 18$ & 80.0 & 96.4 & 97.7 & 64.1 & 88.2 & 93.0 & 519.3\\
	\bottomrule[1.1pt]
	\end{tabular}
	\label{tab:head}
\end{table}

To assess the effect of the number of comb ``teeth'' $(h)$ on modality alignment, we conduct an ablation study (see \cref{tab:head}). Since both the number of regions $N$ and the number of words $L$ are 36 in our experiments, $h$ must be a divisor of 36, \textit{i.e.}, $h \in \{2,3,4,6,9,12,18\}$. 
We find that a large number of combs ($h$) causes each comb to cover only a few tokens (regions or words), which is insufficient to capture phrase-level semantics or composite region patterns. Conversely, a small $h$ forces each comb to compress a large amount of fine-grained information, effectively reducing all tokens to a global representation and weakening fine-grained alignment. Therefore, a balance is needed: we observe that $h = 6$ offers the best trade-off and achieves optimal performance.

\begin{figure*}[!t]
	\centering
	\includegraphics[width = 0.9\linewidth]{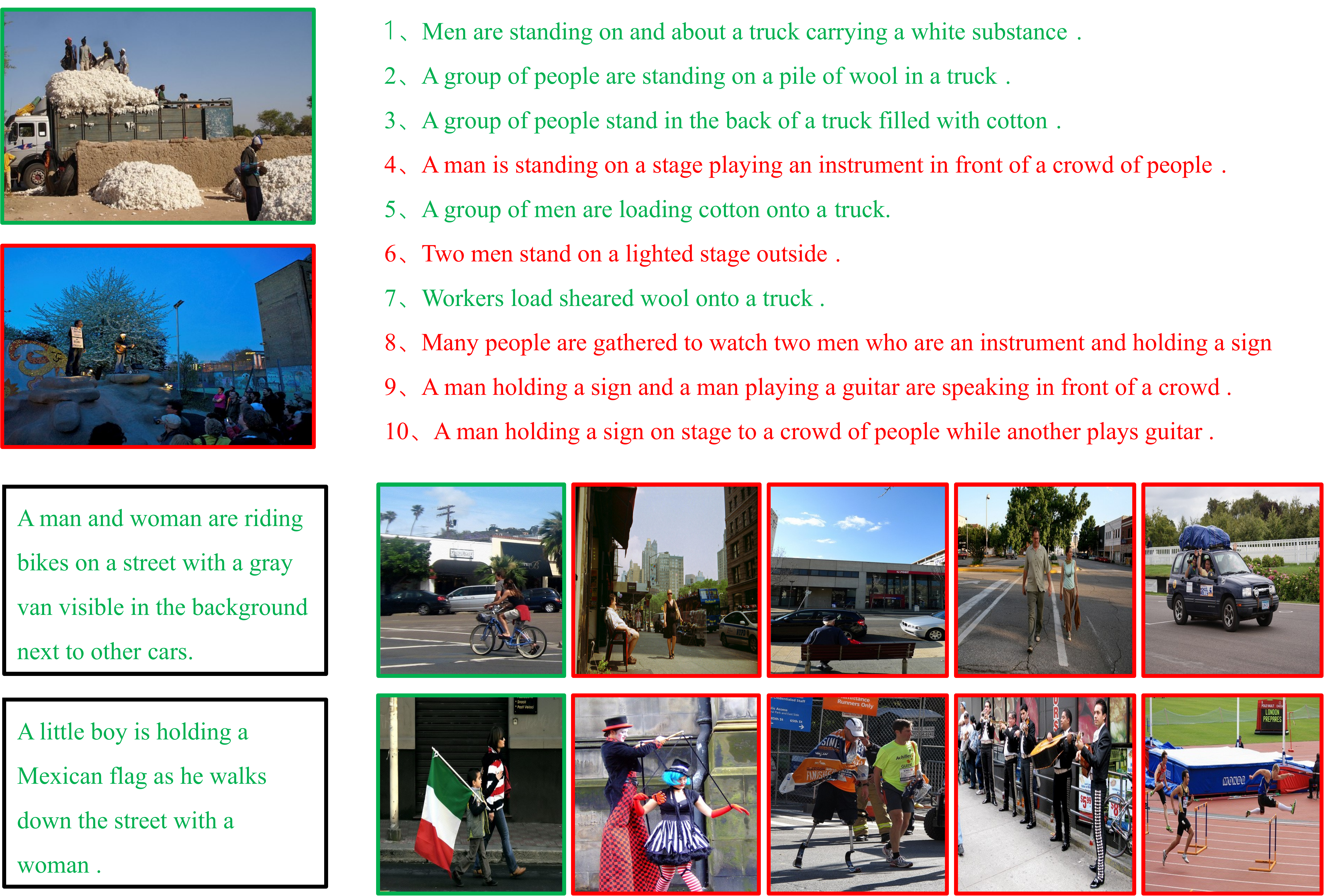} 
	\caption{\textbf{Visualization of retrieval results.} Top: image-to-text retrieval examples. Bottom: text-to-image retrieval examples by CMSF on \textsc{Flickr30K}.}
	\label{fig:result}
\end{figure*}

\section{Visualization and Case Study}

To further demonstrate CMSF’s superiority, we visualize retrieval results on \textsc{Flickr30K} test sets (see \cref{fig:result}). For each image query, we present the top-10 retrieved sentences; for each text query, we show the top-5 retrieved images.
For image queries, CMSF correctly retrieves all relevant sentences. Even when retrieval errors occur, the incorrect sentences come from the same image (highlighted by a red box in the top-left), in both image-text pairs, key words or salient regions such as ``men'', ``group of people'', and ``standing'' appear consistently,
indicating that our model returns images with highly similar scene, content, and composition. This highlights CMSF’s accuracy and robustness in interpreting image content.


For text queries, CMSF consistently ranks the ground-truth image first. In cases where non-ground-truth images appear, the top results still contain objects matching key terms (\textit{e.g.}, ``woman and man'', ``street'', ``gray van'', ``cars''), demonstrating that CMSF reliably captures object-level semantics.

\end{document}